\documentclass[sigconf]{acmart}
\AtBeginDocument{%
  }

\setcopyright{acmlicensed}
\acmYear{2026}
\acmDOI{10.1145/3767308.3835979}
\acmConference[MM '26]{Proceedings of the 34th ACM International Conference on Multimedia}{November 10--14, 2026}{Rio de Janeiro, Brazil}
\usepackage{colortbl}
\usepackage{hyperref}
\usepackage{microtype}
\usepackage{graphicx}
\usepackage{subcaption}
\usepackage{booktabs} % for professional tables
\usepackage{multirow}
\usepackage{algorithm}
\newcommand{\ie}{i.e.,\ }

\begin{document}
\title{MESAM: Enhancing SAM2 for Few-Shot Segmentation \\ through Semantic-Visual Memory Generation and Refinement}

\author{Tianyu Zou}
\email{zoutianyu@whut.edu.cn}
\affiliation{%
  \institution{Sanya Science and Education Innovation Park, Wuhan University of Technology}
  \city{Sanya}
  \state{Hainan}
  \country{China\\}
  \institution{School of Artificial Intelligence, Wuhan University of Technology}
  \city{Wuhan}
  \state{Hubei}
  \country{China}
}
\author{Shengwu Xiong}
\email{xiongsw@whut.edu.cn}
\affiliation{%
  \institution{Hubei Key Laboratory of Transportation Internet of Things, Wuhan University of Technology}
  \city{Wuhan}
  \state{Hubei}
  \country{China}
}
\author{Rui Wu}
\email{wrui@whut.edu.cn}
\affiliation{%
  \institution{School of Artificial Intelligence, Wuhan University of Technology}
  \city{Wuhan}
  \state{Hubei}
  \country{China}
}
\author{Ruilin Yao}
\email{yaoruilin@whut.edu.cn}
\affiliation{%
  \institution{School of Artificial Intelligence, Wuhan University of Technology}
  \city{Wuhan}
  \state{Hubei}
  \country{China\\}
  \institution{Institute of Automation, Chinese Academy of Sciences}
  \city{Beijing}
  \country{China}
}
\author{Yi Rong}
\authornote{Corresponding author}
% \correspondingauthor
\email{yrong@whut.edu.cn}
\affiliation{%
  \institution{Sanya Science and Education Innovation Park, Wuhan University of Technology}
  \city{Sanya}
  \state{Hainan}
  \country{China\\}
  \institution{School of Artificial Intelligence, Wuhan University of Technology}
  \city{Wuhan}
  \state{Hubei}
  \country{China}
}

\begin{abstract}
This paper investigates how to leverage the powerful capabilities of the Segment Anything Model 2 (SAM2) to tackle few-shot segmentation (FSS) tasks. One major challenge lies in the substantial appearance differences between the support and query images, which may diminish the effectiveness of the memory-matching process in SAM2. Through systematic analysis, we find that the memory generated based on the categorical semantic information is robust to this appearance divergence and is more compatible with SAM2. Based on these observations, we propose a new Memory-Enhanced SAM2 (MESAM) method, which first introduces a Semantic-Visual Memory Generation (SVMG) module to separately construct a query semantic memory, a query foreground visual memory, and a support adaptive visual memory. Then, in our designed Interactive Semantic Memory Refinement (ISMR) module, we further enrich the obtained semantic memory with fine-grained visual details, and suppress its misatcivated background responses. Extensive experiments on the PASCAL-$5^i$, COCO-$20^i$, and LVIS-$92^i$ datasets demonstrate that MESAM consistently outperforms existing FSS methods, achieving new state-of-the-art segmentation performance. Our source code is available at: \href{https://github.com/tianyu-zou/MESAM}{https://github.com/tianyu-zou/MESAM}.
\end{abstract}





\maketitle

\section{Introduction}
\label{sec:intro}
Semantic segmentation~\cite{long2015fully, lee2013pseudo, cheng2022masked}  represents a fundamental task in the field of computer vision, having made significant progress in both academic and industrial communities. These achievements are primarily reliant upon the availability of numerous well-annotated training data. However, manually annotating pixel-level object masks for large-scale image samples is laborious and time-consuming. Moreover, traditional segmentation methods struggle to handle new object categories not encountered during training, thus limiting their generalization ability for segmenting arbitrary classes. To overcome these limitations, few-shot segmentation (FSS) \cite{shaban2017one, cheng2022holistic, zou2023segment} has been proposed. By emulating the human intelligence that can understand unseen concepts through observing only a few examples, FSS attempts to utilize transferable prior knowledge learned from some base classes, alongside useful information extracted from limited annotated support samples, to segment query images containing target objects belonging to novel categories. However, due to the absence of novel class data during training, most existing FSS models tend to overfit to base classes, thereby diminishing the transferability of their learned prior knowledge.

\begin{figure}[t]
    \centering
    \includegraphics[width=0.9\linewidth]{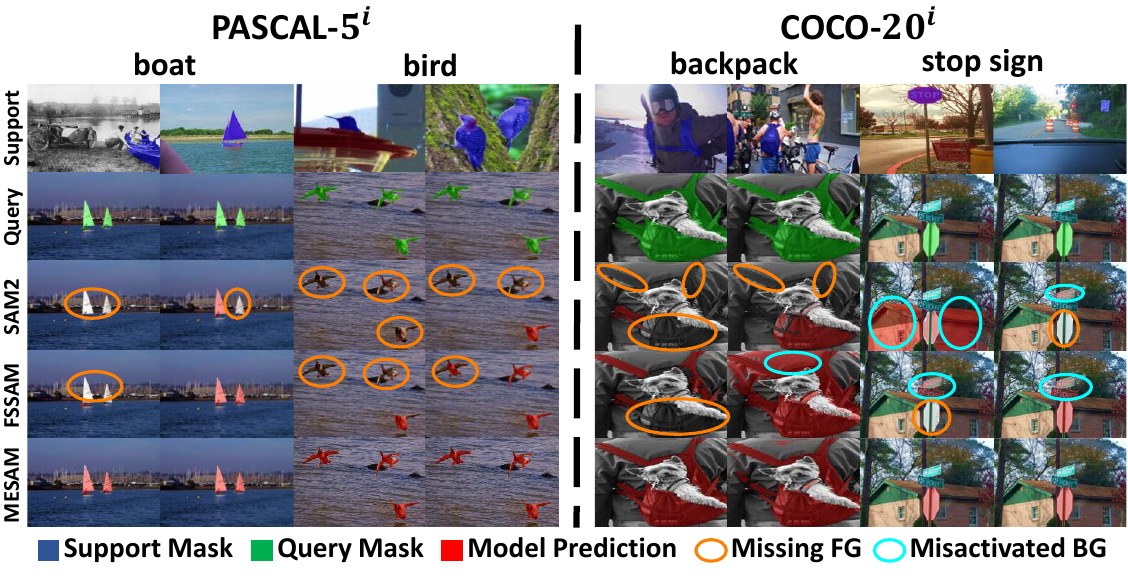}
    \caption{Illustration of the large support-query appearance divergence challenge. With only partial visual cues provided by the support image, existing methods segment only visually similar parts in the query, instead of the entire object. In contrast, with the refined semantic memory, our MESAM method yields better segmentation results. The orange and blue circles indicate foreground (FG) misses and background (BG) misactivations, respectively.}
    \label{fig:Intra_class}
    \vspace{-10pt}
\end{figure}

To address this problem, some works~\cite{zhou2024unlocking, mao2025one, meng2024segic} have begun to explore the use of vision foundation models (e.g., DINOv2~\cite{oquab2023dinov2} and SAM~\cite{kirillov2023segment}) to leverage their fine-grained visual representations or prompt-based segmentation capability. These will provide more transferable prior knowledge or more effective adaptation procedures, thus enhancing the learning of FSS models. More recently, SAM2~\cite{ravi2024sam} has demonstrated impressive inter-frame foreground-to-foreground (FG-FG) matching ability in tackling video segmentation tasks, which may also be beneficial for solving FSS tasks. However, unlike objects exhibiting high visual similarity across consecutive video frames, \textbf{the targets in support and query images within an FSS task typically present significant variations in appearance}. This will prevent the direct use of SAM2 in FSS, since a substantial misalignment will exist between the query samples and the SAM2 segmentation memory generated from support images, resulting in performance degradation (see the third row in Fig.~\ref{fig:Intra_class}). 

Despite emerging studies \cite{zhang2023tale, zhang2024telling, bai2024revsam2} attempting to reduce the visual difference between support and query images, such as by generating intermediate samples~\cite{wang2025correspondence}, predicting query-oriented prompts~\cite{xu2025unlocking}, or using feature adaptation~\cite{cuttano2025sansa}, the large support-query appearance divergence remains an unsolved challenge when applying SAM2 to FSS. As illustrated in Fig.~\ref{fig:Intra_class}, when the support sample presents only partial appearances of the target class (e.g., showing the hull of a boat without sail or only the shoulder straps of a backpack), even the state-of-the-art SAM2-based FSSAM method~\cite{xu2025unlocking} can only segment the visually similar object parts in the query image, rather than the entire object. We analyze that it is because \textbf{current methods construct SAM2 memory based solely on visual cues, but support images alone cannot provide the essential evidence required for accurate query segmentation, due to their visual discrepancies.}

In this paper, we explore a more effective memory mechanism to enhance the capability of SAM2 to address the above challenge in FSS. We argue that utilizing the categorical textual descriptions is a better choice for memory construction, since they can introduce class-specific semantic information that is not affected by support-query appearance divergence. We conduct two toy experiments to show the superiority of the semantic memory and explain why it is more compatible with SAM2. Based on this analysis, we propose a \textbf{M}emory-\textbf{E}nhanced \textbf{SAM}2 (MESAM) method for FSS, which exploits semantic and visual information in a complementary manner. Specifically, our method first develops a Semantic-Visual Memory Generation (SVMG) module to produce a query semantic memory based on the GradCAM~\cite{selvaraju2017grad} of CLIP's~\cite{radford2021learning} text classifier, whilst simultaneously generating a query foreground (FG) visual memory and a support adaptive visual memory based on the DINOv2 image representations. Then, in the following Interactive Semantic Memory Refinement (ISMR) module, the query semantic memory is refined by interacting with the query FG memory to absorb its useful visual detail information. Additionally, the support adaptive memory is employed to further calibrate this refinement process, suppressing background (BG) responses and preventing excessive activation from visual cues. \textbf{In this way, the obtained refined memory representations integrate both semantic information that is robust to support-query appearance divergence and fine-grained visual features that are beneficial to precise segmentation} (see the bottom row in Fig~\ref{fig:Intra_class}). Finally, this refined semantic memory is passed through SAM2's memory attention and mask decoder to generate the final segmentation mask. 

\begin{figure*}[t]
    \centering
    \includegraphics[width=0.8\linewidth]{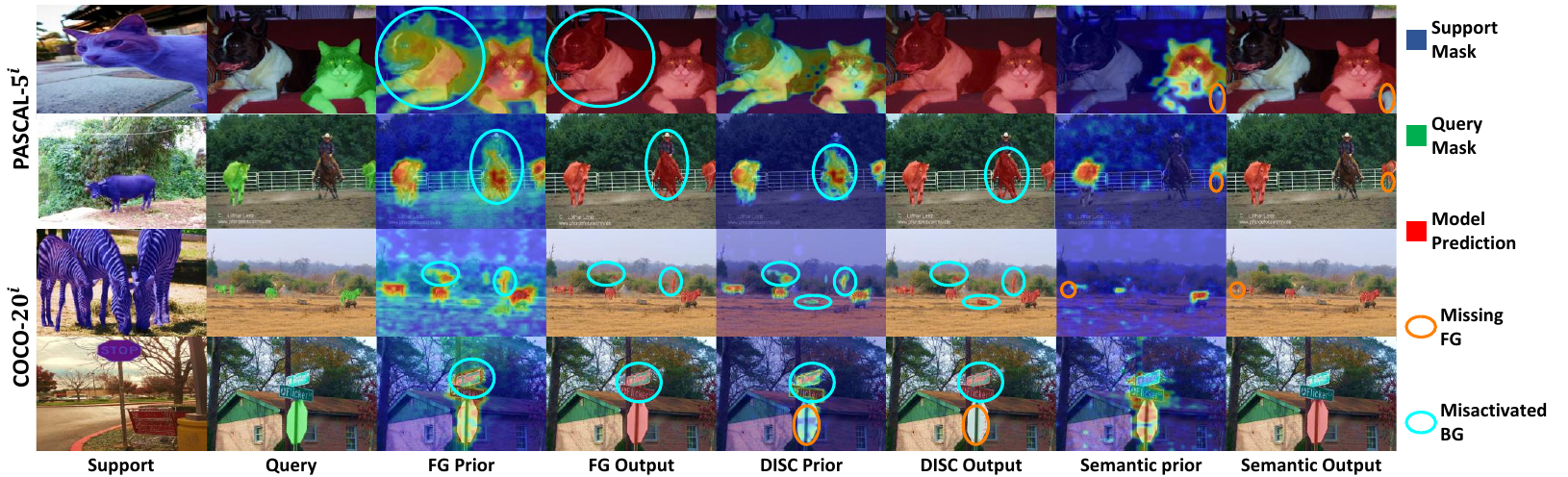}
    \caption{The priors and the segmentation results of two visual memories (\ie FG, DISC) and our semantic memory.}
    \label{fig:visual_semantic}
\end{figure*}

Our main contributions are threefold: 1) We discover that the memory generated from categorical semantic information is more compatible with SAM2 and leads to better FSS performance, which motivates the design of a memory mechanism that integrates both robust semantic cues and visual details. 2) We propose a new SAM2-based MESAM method for FSS and design two modules, \ie SVMG and ISMR, for constructing and refining memory representations, respectively. Based on our knowledge, MESAM is the first attempt to exploit semantic information to empower SAM2 for mitigating the challenge of large support-query appearance divergence in FSS. 3) Extensive experiments on the PASCAL-$5^i$, COCO-$20^i$, and LVIS-$92^i$ datasets show that MESAM outperforms existing state-of-the-art methods, validating the effectiveness of our designed semantic-visual memory mechanism.

\vspace{-3pt}
\section{Related work}
\label{sec:related_work}
\noindent {\textbf{Traditional Few-Shot Segmentation.}} 
Early FSS methods are dominated by prototype learning~\cite{liu2022dynamic,hariharan2011semantic,zhang2021self}, evolving from single-prototype models~\cite{zhang2020sg} to multi-part and query-related prototypes~\cite{yang2020prototype,li2021adaptive,fan2022self,peng2023hierarchical}. To alleviate the detail loss of prototype compression, affinity learning~\cite{hu2019attention,zhang2022mask,wang2020few,xie2021scale} constructs dense support–query correlations (e.g., cycle correlation~\cite{zhang2021few}, hyper correlation~\cite{min2021hypercorrelation,shi2022dense}). Such affinity methods often suffer from FG-BG misalignment, motivating self-calibration and feature purification~\cite{xu2023self, xu2025eliminating, zou2025balancing}. Despite these advances, many FSS models may still overfit to base classes, limiting the transferability of learned priors to unseen concepts.

\noindent {\textbf{Foundation Based Few-Shot Segmentation.}} 
Vision foundation models such as DINOv2~\cite{oquab2023dinov2} and SAM~\cite{kirillov2023segment} provide rich semantic priors and strong generalization, addressing limitations of traditional FSS. Training-free methods like Matcher~\cite{ICLR2024_4df9a5e6} and GF-SAM~\cite{zhang2024bridge} use DINOv2 features to prompt SAM, while VRP-SAM~\cite{sun2024vrp}, MM-Promptt~\cite{xiong2025mm}, and ProSAM~\cite{wang2025prosam} design prompt encoder for frozen SAM models. SAM2~\cite{ravi2024sam} introduces a memory mechanism for same-object FG–FG matching, but large appearance differences between support and query images in FSS can cause memory misalignment and degraded performance~\cite{zu2025rethinking, xing2025sam2}. Recent adaptation strategies include FSSAM~\cite{xu2025unlocking} introduces external visual guidance from DINOv2, SANSA~\cite{cuttano2025sansa} leverages the latent semantic knowledge in SAM2, and CAV-SAM~\cite{wang2025correspondence} uses diffusion-generated intermediate frames. Despite these efforts, the large appearance differences between support and query images remains an unresolved challenge when applying SAM2 to FSS. Different from all existing methods, our proposed MESAM attempts to use semantic information to construct more effective and robust memory representations for SAM2 so as to address this challenge.

\begin{table}[t]
\vspace{-10pt}
    \centering
    \caption{Performance comparison between the visual memories constructed in FSSAM and our semantic memory.}
    \vspace{-2pt}
    \resizebox{0.6\linewidth}{!}{
        \begin{tabular}{lccc}
        \toprule
            Memory & FG & DISC & Semantic \\ \midrule
            PASCAL-$5^i$ & 81.1 & 78.1 & 82.8  \\
            COCO-$20^i$ & 55.9 & 56.0 & 58.9  \\
            \bottomrule
        \end{tabular}
    }
    \label{tab:visual_semantic}
    \vspace{-10pt}
\end{table}

\begin{table}[t]
    \centering
    \caption{The false positive (FP) and false negative (FN) rates of the prior masks used for generating FG and DISC visual memories, and our semantic memory.}
    \vspace{-2pt}
    \resizebox{0.7\linewidth}{!}{
        \begin{tabular}{l|cc|cc}
        \toprule
            & \multicolumn{2}{c|}{PASCAL-$5^i$} & \multicolumn{2}{c}{COCO-$20^i$}\\
            Memory & FP & FN & FP & FN \\ \midrule
            FG & 0.0865  & 0.0436 & 0.1271  & 0.0381  \\
            DISC & 0.0186 & 0.1056 & 0.0324 & 0.0820 \\
            Semantic & 0.0139 & 0.1207 & 0.0155 & 0.0915   \\
            \bottomrule
        \end{tabular}
    }
    \label{tab:FP_FN}
    \vspace{-10pt}
\end{table}

\section{Methodology}
\label{sec:method}
The main goal of this work is to learn a segmentation model on a training set $\mathcal{D}_{\text{train}}$ with sufficient data from base classes $\mathcal{C}_{\text{base}}$, expecting that the learned model can generalize well on the FSS tasks sampled from a testing set $\mathcal{D}_{\text{test}}$ of novel classes $\mathcal{C}_{\text{novel}}$, which are not overlapped with $\mathcal{C}_{\text{base}}$, e.g., $\mathcal{C}_{\text{base}} \cap \mathcal{C}_{\text{novel}} = \emptyset$. Following the standard setup, we describe the $k$-shot segmentation task as an episode  $\mathcal{E}=\{\mathcal{S}, \mathcal{Q}\}$ that consists of a support set $\mathcal{S} = \{(I_s^{n}, M_s^{n})\}_{n=1}^k$ with $k$ samples for model adaptation and a query set $\mathcal{Q} = (I_q, M_q)$ for performance evaluation. Here, $\mathcal{S}$ and $\mathcal{Q}$ are two disjoint sets drawn from the same specific object category, satisfying $\mathcal{S} \cap \mathcal{Q} = \emptyset$. $I \in \mathbb{R}^{H \times W \times 3}$ and $M \in \{0,1\}^{H \times W}$ denote an input image and its binary segmentation mask, where $H$ and $W$ indicate their height and width, respectively. To deal with testing FFS tasks, an episodic learning strategy is usually employed. It constructs a series of episodes from $\mathcal{D}_{\text{train}}$ for model training to simulate the processing of the FFS tasks that will be encountered at the evaluation stage.

\vspace{-2pt}
\subsection{Analysis of Semantic Memory}
\label{Analysis_of_Semantic_Memory}

In this subsection, we investigate two questions: 1) Whether generating the semantic memory for SAM2 based on the categorical texts yields better FSS segmentation results? 2) If so, why such semantic memory is more compatible with SAM2 when processing FSS tasks? Following FSSAM~\cite{xu2025unlocking}, we construct our visual memory based on the prior mask generated for the query image. Semantic memory is derived from the GradCAM of CLIP's text classifier, which categorizes query images into FG and BG classes with and without the target object. The details of this construction process will be presented in the following subsections.

For the first question, we compare our semantic memory with the FG and DISC visual memories constructed by FSSAM to validate its effectiveness. We present their segmentation performance measured by mIoU in Table~\ref{tab:visual_semantic}. Their segmented results and the prior masks used for generating these memories are also visualized in Fig.~\ref{fig:visual_semantic}. As can be seen, since the prior masks for the FG and DISC memories are calculated from the visual similarity between the support prototypes and query image pixels, these prototypes may incorrectly activate some visually similar BG areas in query samples (e.g, the dog and the horse in the first two rows of Fig.~\ref{fig:visual_semantic}), while overlooking FG objects with great appearance differences (e.g, the stop sign in the bottom row of Fig.~\ref{fig:visual_semantic}), thus leading to relatively lower segmentation accuracy. By contrast, the GradCAM of CLIP's text classifier captures the semantic information of the target class, allowing the prior mask for constructing our semantic memory to be unaffected by the visual representations of support images. Consequently, this memory can produce more reliable predictions and obtain overall higher mIoU values (82.8\% and 58.9\% on the two datasets), verifying its superiority.

\begin{figure}[t]
    \centering
    \includegraphics[width=0.8\linewidth]{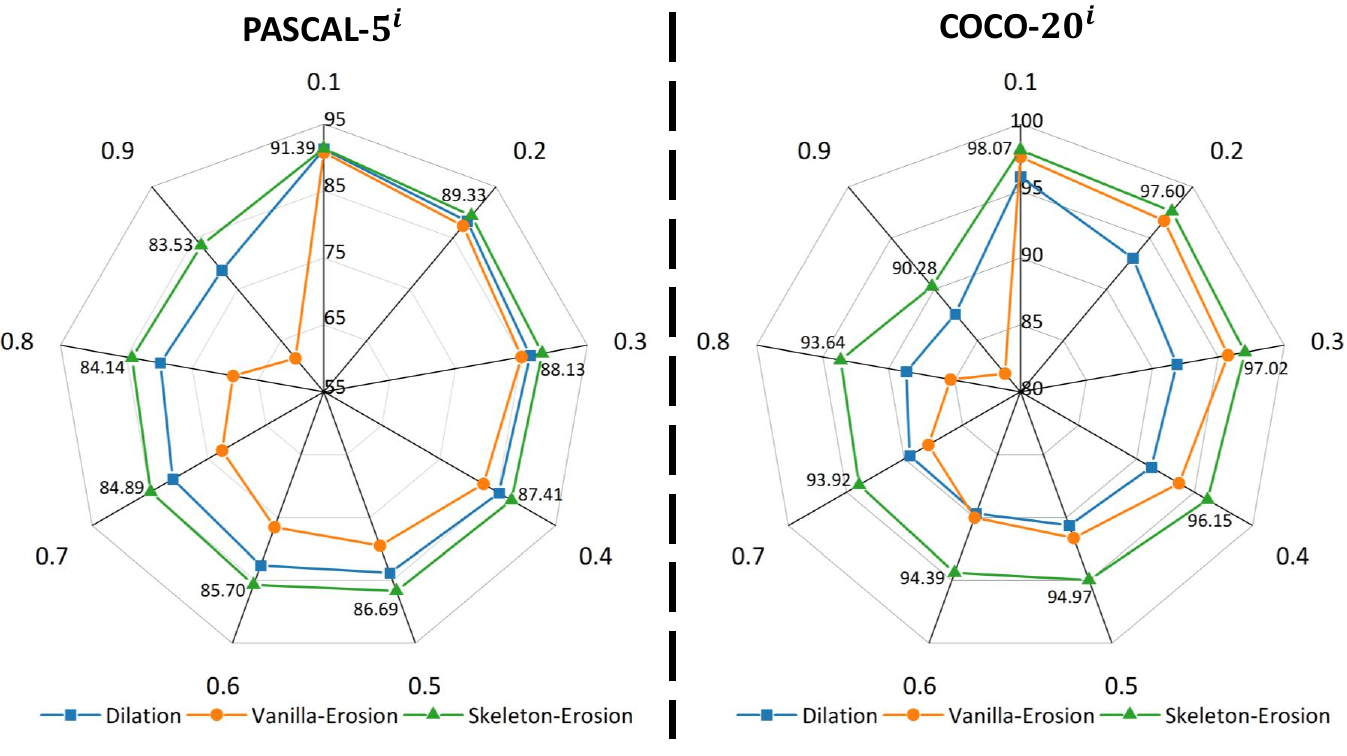}
    \vspace{-3pt}
    \caption{Performance comparison between different memories obtained under varying dilation and erosion ratios.}
    \label{fig:dilate_erode}
    \vspace{-12pt}
\end{figure}

Next, we attempt to answer the second question in order to explain the effectiveness of semantic memory. To do this, we first convert the prior masks for each memory on different query images into binary masks using a threshold of 0.5, and then report their average false positive (FP) and false negative (FN) rates in Table~\ref{tab:FP_FN}. We can find that the prior masks for visual memories typically have higher FP but lower FN rates, indicating that they typically cover more FG pixels and also incorrectly over-segment certain BG areas. In contrast, the prior masks for our semantic memory tend to only cover reliable pixels that are strongly identified as belonging to FG areas, thus reducing FP rates while increasing FN rates. Based on these statistics, we conduct a toy experiment, which applies dilation and erosion operations to the ground-truth segmentation masks of query samples to simulate the prior masks used for generating the visual and semantic memories, respectively. Note that, in addition to the standard erosion operation (\ie vanilla-erosion), we also employ an enhanced erosion process that preserves the skeleton of target objects unchanged (\ie skeleton-erosion).

\begin{figure*}[t]
    \centering
    \includegraphics[width=0.8\linewidth]{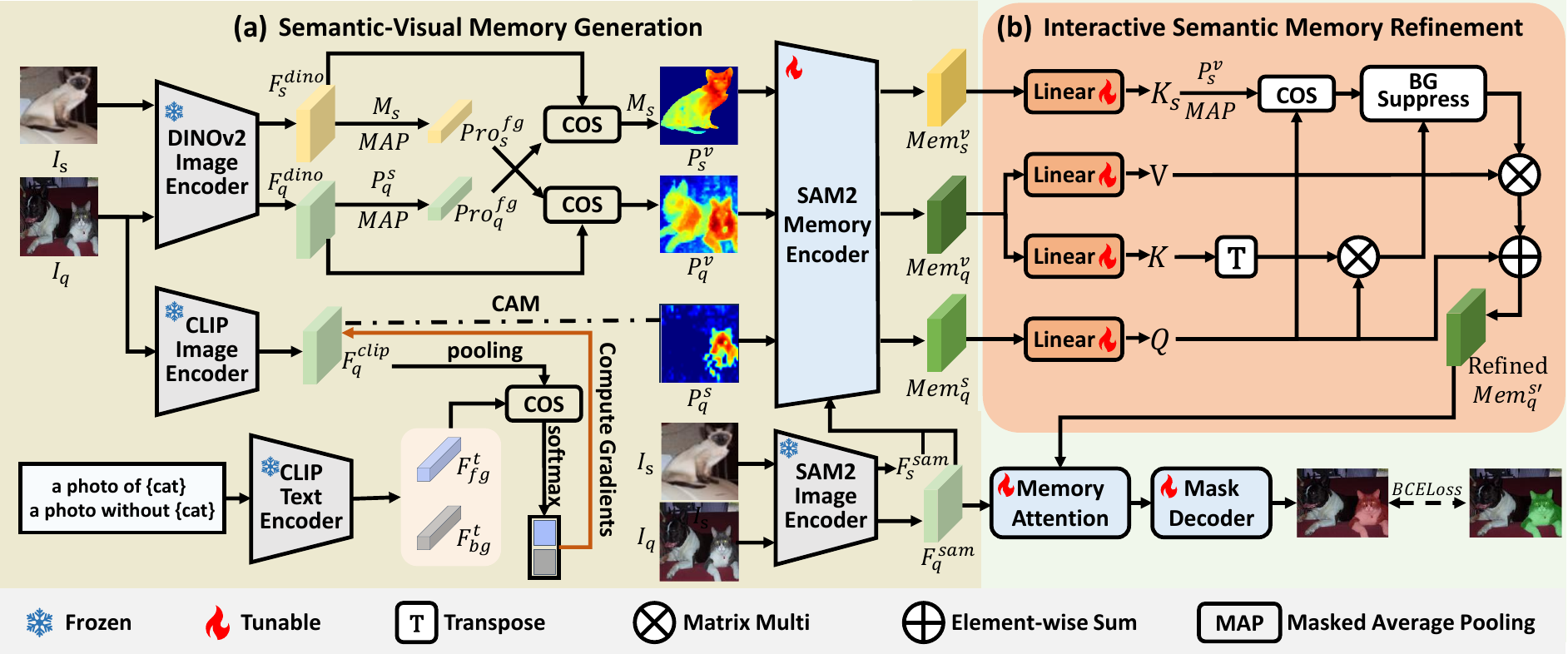}
    \caption{Illustration of our MESAM method. (a) The SVMG module generates three initial memories $Mem^s_q$, $Mem^v_q$, and $Mem^v_s$. (b) In the ISMR module, $Mem^s_q$ is enhanced by interacting with $Mem^v_q$ through the cross-attention operation, while $Mem^v_s$ is used to suppress the BG noise in the attention scores. Subsequently, the refined semantic memory ${Mem_q^s}'$ is processed by SAM2's memory attention module and mask decoder to produce the final segmentation mask.}
    \label{fig:overall_framework}
    \vspace{-10pt}
\end{figure*}

Fig.~\ref{fig:dilate_erode} displays the segmentation results achieved by different memories constructed from these prior masks under varying dilation and erosion rates. We can observe that: 1) The skeleton-erosion memory consistently outperforms the dilation memory, suggesting that erroneously introducing BG noise like the visual memories may impose a more severe negative impact on SAM2 segmentation than the loss of some FG information. 2) The vanilla-erosion memory without retaining the target object's skeleton results in performance degradation compared to the skeleton-erosion memory and even the dilation memory. This indicates that preserving key FG features is also critical for SAM2 to achieve precise segmentation. Based on these observations, we can illustrate the compatibility between our semantic memory and SAM2: Since the semantic information introduced by the CLIP classifier is classification-oriented global information shared across the entire target category, it exhibits high \textbf{reliability} and \textbf{representativeness}. It is less likely to introduce BG noise and will capture representative object features that are essential for classification. Therefore, with these properties, our semantic memory can work well with SAM2 for better FSS results.

\vspace{-3pt}
\subsection{Our Proposed MESAM Method} \label{sec:method_pipline}
Based on the above analysis, we build our MESAM framework primarily upon the semantic memory. However, since the global categorical semantic information cannot fully represent all FG features, utilizing only the semantic memory will result in insufficient segmentation of several target objects in some query images (see orange circles in Fig.~\ref{fig:visual_semantic}). To address this issue, we further integrate the visual information into our semantic memory, allowing it to be refined through the incorporation of additional FG details. Next, we will detail how we achieve this through the two modules we designed, namely SVMG and ISMR. For clarity, this introduction will be presented under the 1-shot ($k=1$) setting, yet it can be easily extended to the scenarios of $k > 1$.

\noindent \textbf{Semantic-Visual Memory Generation (SVMG).} As illustrated in Fig.~\ref{fig:overall_framework}(a), SVMG aims to construct three types of memories, including a query semantic memory, a query FG visual memory, and a support adaptive visual memory. To this end, SVMG first calculates corresponding prior masks to generate these memories as follows.

\textit{1) Query Semantic Prior Mask:} For each object category to be segmented, we define the texts describing its FG and BG concepts using the prompts ``a photo of \{target class\}'' and ``a photo without \{target class\}'', respectively. The CLIP text encoder takes these descriptions as input and produces the weights $F^t_{fg}$ and $F^t_{bg}$ for the FG and BG classes of a binary text classifier. Given a query image $I_q$, we can get its feature maps $F^{clip}_q\in\mathbb{R}^{h \times w \times d}$ through the CLIP image encoder, where $d$ denotes the number of channels, $h$ and $w$ represent the height and width, respectively. We then apply the representative average pooling on $F^{clip}_q$ to obtain the global embedding $f^{clip}_q \in \mathbb{R}^{d}$ of $I_q$, and classify it into the FG and BG classes based on the following classification scores:
\begin{equation}
    S_{i}=\frac{\text{exp}(\tau\text{Cos}(f^{clip}_q, F^t_{i}))}{\Sigma_{j\in\{fg,bg\}}\text{exp}(\tau\text{Cos}(f^{clip}_q, F^t_{j}))}, i \in \{fg, bg\}.
\end{equation}
Here, $\text{Cos}(\cdot, \cdot)$ denotes cosine similarity, $\tau$ is a positive temperature coefficient. Following GradCAM, we calculate the gradient of the FG classification score $S_{fg}$ with respect to each feature map of $F^{clip}_q$:
\begin{equation} \label{eq:relevance_weight}
    w_m = \frac{1}{hw} \sum_i \sum_j \frac{\partial S_{fg}}{\partial{F^{clip}_{q,m}}(i,j)},
\end{equation}
where $F^{clip}_{q,m}$ denotes the $m$-th feature map, and $(i,j)$ indexes the spatial location of each pixel. According to this definition, $w_m$ represents the semantic relevance between the $m$-th feature map and the FG class. By taking $w_m$ as the weight, we aggregate all feature maps to obtain our query semantic prior mask $P^s_q$ as follows:
\begin{equation} \label{eq:query_semantic}
    P_q^s = \text{ReLU}(\sum_m w_m F^{clip}_{q,m}),
\end{equation}
where $\text{ReLU}(\cdot)$ filters out negative responses. In this way, $P^s_q$ highlights the image regions within $I_q$ that make greater contributions towards classifying $I_q$ into the FG class, which are highly relevant to the target object semantics.

\textit{2) Query FG Visual Prior Mask:} To refine the semantic memory with complementary FG details, we need to extract FG-related visual information from the query image. For this purpose, we first employ the DINOv2 image encoder to convert the support and query images into their visual features $F^{dino}_s$ and $F^{dino}_q$, respectively. Then, based on the ground-truth mask $M_s$ of the support image $I_s$, we perform the masked average pooling (MAP) operation on $F^{dino}_s$ to generate the support FG prototype $Pro^{fg}_s$. This prototype is subsequently utilized as an indicator for identifying the FG visual features in $F^{dino}_q$, so that our query FG visual prior mask $P^v_q$ can be generated as follows:
\begin{equation} \label{eq:visual_priors}
    P^v_q = \text{Norm}(\text{Cos}(F^{dino}_q, Pro^{fg}_s)).
\end{equation}
Here, $\text{Norm}(\cdot)$ denotes the min-max normalization. By doing so, $P^v_q$ emphasizes the query image regions that exhibit more consistent features with the support FG areas.

\textit{3) Support Adaptive Visual Prior Mask:} We also exploit the visual features of the support image to enhance our semantic memory. Unlike previous works that directly treat $M_s$ as the support prior mask, we attempt to leverage the query-oriented FG information in $F^{dino}_s$ to generate an adaptive prior mask, thereby further mitigating the support-query appearance divergence problem. Concretely, we first apply the MAP operation to $F^{dino}_q$ to compute the query FG prototype $Pro^{fg}_q$ according to the semantic prior mask $P^v_q$ defined in Eq.~(\ref{eq:query_semantic}). Then, by comparing $Pro^{fg}_q$ with $F^{dino}_s$, we obtain the support adaptive visual prior mask $P^v_s$ as follows:
\begin{equation}
    P^v_s = M_s \odot \text{Norm}(\text{Cos}(F^{dino}_q, Pro^{fg}_s)).
\end{equation}
Here, $\odot$ represents the element-wise multiplication, and $M_s$ is used to filter out the responses located in the BG regions. With this definition, $P^v_s$ highlights partial FG areas within $I_s$ that are more similar to the FG regions of $I_q$.

After obtaining the three prior masks $P^s_q$, $P^v_q$, and $P^v_s$, we interpolate them to the size of input images $H\times W$, and then feed them into the SAM2 memory encoder $\text{MemEnc}(\cdot, \cdot)$ to construct the corresponding memories:
\begin{equation}
\begin{aligned}
    Mem^s_q &= \text{MemEnc}(F_q^{sam}, P^s_q), \\
    Mem^v_q &= \text{MemEnc}(F_q^{sam}, P^v_q), \\
    Mem^v_s &= \text{MemEnc}(F_s^{sam}, P^v_s), \\
\end{aligned}
\end{equation}
where $F_s^{sam}$ and $F_q^{sam}$ denote the support and query image features produced by the SAM2 image encoder. These three memories share the same size of $h_mw_m \times d_m$. Here, $h_mw_m$ denotes the spatial size and $d_m$ is the number of dimensions. 

% \subsection{Overview of Our MESAM}
% To extract effective memory for FSS, we build the MESAM model by integrating semantic memory from text-driven activation and visual memory from support–query correspondence to supply fine-grained details that semantic matching alone may miss, as illustrated in Figure~\ref{fig:overall_framework}. For memory construction, we first generate a semantic prompt $P_q^t$ on the query image $I_q$ using a text prompt with GradCAM, and a visual prompt $P_q^v$ by comparing the support foreground prototype $Pro_s^f$ with the query DINOv2 features $F_q^{\text{dino}}$. Similarly, the support visual prompt $P_s^v$ is obtained by comparing the query foreground prototype $Pro_q^f$ with the support features $F_s^{\text{dino}}$. All prompts are then fused with their corresponding SAM2 visual features ($F_q^{sam}$ for the query, $F_s^{sam}$ for the support) and encoded by the SAM2 memory encoder into semantic and visual memories $\{Mem_q^t, Mem_q^v, Mem_s^v\}$. The semantic memory $Mem_q^t$ and the visual memories $Mem_q^v$, $Mem_s^v$ are sequentially processed by Support-Calibrated Memory Refinement (SCMR) module, which leverages the $Mem_q^v$ to enhance foreground consistency and suppress background responses by $Mem_s^v$, producing a robust memory representation $\hat{Mem}_q^t$ for the query image. The effective memory $\hat{Mem}_q^t$ and query features $F_q^{sam}$ are fed into Dense Feature Matching module to produce the final segmentation mask $\hat{y}$.

\noindent \textbf{Interactive Semantic Memory Refinement (ISMR).} In this module, the query semantic memory $Mem^s_q$ is further refined by absorbing fine-grained visual information from $Mem^v_q$ through a cross-attention operation. However, since some BG areas in $P_q^v$ may be over-activated by the similar visual cues in $Pro^{fg}_s$, $Mem^v_q$ will inevitably introduce BG noise that affects such refinement process. To address this problem, considering that $M_s$ has already removed the BG responses from $P^v_s$, the resulting $Mem^v_s$ contains only pure FG information. We therefore use it to suppress BG noise in the cross-attention scores generated above.

Specifically, $Mem^s_q$, $Mem^v_q$, and $Mem^v_s$ are first projected into a $d_c$-dimensional common space:
\begin{equation}
\begin{aligned}
Q &= \text{Linear}_Q(Mem^s_q), & &K= \text{Linear}_K(Mem^v_q), \\
V &= \text{Linear}_V(Mem^v_q), & &K_s= \text{Linear}_{K_s}(Mem^v_s),
\end{aligned}
\end{equation}
where $\text{Linear}_i(\cdot)$ indicate the linear projectors with different parameters. The resulting representations $\{Q, K, V, K_s\} \in \mathbb{R}^{N \times d_c}$ ($N=h_mw_m$) will be used for subsequent semantic memory refinement. Following the cross-attention mechanism, we calculate the scaled dot-product attention scores ${A}_{qq} \in \mathbb{R}^{N \times N}$ between two query memories $Q$ and $K$:
\begin{equation}
    {A}_{qq} = {Q \cdot K^\top}/{\sqrt{d_c}}.
\end{equation}
To handle the BG noise in ${A}_{qq}$, the MAP operation is first imposed on $K_s$ and $P^v_s$ to produce a query-oriented support FG prototype ${Pro}_s \in \mathbb{R}^{1\times N}$. We then use it to scan each spatial location of $Q$ through the following cosine similarity, to identify potential BG regions in the query memories:
\begin{equation} \label{eq:cosine_similarity}
    {A}^i_{sq} = \frac{{Pro}_s \cdot Q_i^\top}{||{Pro}_s||\cdot||Q_i||}, \quad i=1,2,...,N
\end{equation}
where $Q_i$ denotes the $i$-th feature vector in $Q$. The entry ${A}^i_{sq} \in [-1, 1]$ with a smaller value indicates the pixel that is highly likely to be located within the BG regions, as it is less similar to ${Pro}_s$. Note that in Eq.~(\ref{eq:cosine_similarity}), we use $Q$ rather than $K$ to calculate ${A}_{sq} \in [-1, 1]^{1\times N}$ because the semantic memory $Mem^s_q$ typically yields more reliable predictions than $Mem^v_q$, which may benefit the identification of BG pixels. Next, by normalizing the values of ${A}_{qq}$ and ${A}_{sq}$ into $[0,1]$, we obtain the following attention calibration scores:
\begin{equation} \label{eq:calibration}
    A_{qq}' = \text{min}[\text{Norm}({A}_{qq}) + (\text{Norm}({A}_{sq}) - 1), 0].
\end{equation}
Here, $\text{min}[\cdot,0]$ preserves the negative entries in $A_{qq}'$ to suppress the BG activations. Finally, we calibrate the attention scores ${A}_{qq}$ and get the refined semantic memory as follows:
\begin{equation}
    A_{qq}'' = {A}_{qq} + \alpha A_{qq}',
\end{equation}
\begin{equation}
    {Mem_q^s}' = {Mem}_q^s + \text{Linear}_O(\text{Softmax}(A_{qq}'') \cdot V),
\end{equation}
where $\text{Softmax}(\cdot)$ denotes the softmax computation along the column. $\alpha$ is a scaling factor that is set to 10.
% 要说清楚为什么要跟Q算cos而不跟K算cos，通过可视化图来说明，Q是一个保守的prompt得到的，而K是通过激进的prompt得到的，这意味着K激活的前景区域包含一些背景，且这些背景本身就是由support激活的，从而导致K中的背景区域更容易与Ks产生激活，从而在后面的背景抑制中漏了一些背景；而Q激活区域的背景噪声较少，且是通过semantic text激活的，Q与Ks的高响应区域应该为前景，从而在背景抑制中能全面抑制背景

\noindent \textbf{Segmentation Mask Prediction.}
Following the standard pipeline of SAM2, the refined semantic memory ${Mem_q^s}'$ is jointly input with $F_q^{sam}$ into a memory attention module to obtain their matched query features. These features are subsequently decoded by SAM2's mask decoder to predict the final segmentation mask $\hat{M}_q$.

\noindent \textbf{Training Loss.}
We adopt the Binary Cross-Entropy (BCE) loss between the predicted and ground-truth masks to optimize our MESAM model, which is formulated as:
\begin{equation}
\mathcal{L}_{\text{BCE}} = -[ M_q \log(\hat{M}_q) + (1 - M_q)\log(1 - \hat{M}_q)].
\end{equation}

\begin{table*}[t]
    \centering
    \caption{Performance comparisons on PASCAL-$5^i$. ``$5^i$'' shows the mIoU scores of 5 novel classes in fold $i$, ``Mean'' denotes the averaged mIoU score across all four folds. The best and second-best results are highlighted in \textbf{bold} and \underline{underlined}, respectively.}
    \vspace{-2pt}
    \resizebox{0.8\textwidth}{!}{
    \begin{tabular}{l|c|c|cccccc|cccccc}
        \toprule
        & & & \multicolumn{6}{c|}{1-shot} & \multicolumn{6}{c}{5-shot} \\
        Methods & Backbone & Venue & $5^0$ & $5^1$ & $5^2$ & $5^3$ & Mean & FB-IoU & $5^0$ & $5^1$ & $5^2$ & $5^3$ & Mean & FB-IoU \\ \midrule
        ABCBNet~\cite{zhu2024addressing} & ResNet50 & CVPR’24 & 72.9  & 76.0  & 69.5  & 64.0  & 70.6 & -  & 74.4  & 78.0  & 73.9  & 68.3  & 73.6 & - \\
        PAHNet~\cite{zou2025balancing} & ResNet50 & ICCV‘25 & 72.1  & 76.4  & 71.0  & 66.6  & 71.5 & 82.4 & 75.8  & 78.6  & 76.9  & 73.2  & 76.1 & 85.7  \\
        OCNet~\cite{wen2025object} & ResNet50 & ICCV‘25 & 73.5  & 75.9  & 71.1  & 64.9  & 71.4 & 82.2 & 75.9  & 77.1  & 74.1  & 70.9  & 74.5 & 84.7 \\ \midrule
        Matcher~\cite{ICLR2024_4df9a5e6} & SAM-H & ICLR'24 & 67.7  & 70.7  & 66.9  & 67.0  & 68.1 & -  & 71.4  & 77.5  & 74.1  & 72.8  & 74.0 & -  \\
        VRP-SAM~\cite{sun2024vrp} & SAM-H & CVPR'24 & 73.9  & 78.3  & 70.6  & 65.1  & 71.9 & - & - & - & - & - & - & - \\
        GF-SAM~\cite{zhang2024bridge} & SAM-H & NIPS’24 & 71.1  & 75.7  & 69.2  & 73.3  & 72.1 & - & 81.5  & 86.3  & 79.7  & 82.9  & 82.6 & - \\
        FCP~\cite{park2025foreground} & SAM-H & AAAI'25 & 74.9  & 77.4  & 71.8  & 68.8  & 73.2 & - & 77.2  & 78.8  & 72.2  & 67.7  & 74.0 & - \\
        FounFSS~\cite{chang2024high} & DINOV2 & Arxiv'24 & 76.5  & 81.3  & 72.1  & 77.4  & 76.8 & - & 79.5  & 84.8  & 75.8  & 82.5  & 80.7 & - \\
        FS-DINO~\cite{zhuo2025dinov2} & DINOV2 & Arxiv'25 & 75.6  & 80.3  & 72.2  & 73.4  & 75.4 & - & 78.1  & 83.1  & 73.6  & 76.3  & 77.8 & - \\
        FS-SAM2~\cite{forni2025fs} & SAM2-B & ICIP'25 & 74.0  & 79.6  & 73.5  & 66.7  & 73.4 & -  & 75.7  & 79.6  & 75.3  & 68.8  & 74.8 & -  \\
        FSSAM~\cite{xu2025unlocking} & SAM2-S & ICML'25 & \underline{81.6}  & \underline{84.9}  & \underline{81.6}  & \underline{76.0}  & \underline{81.0} & \underline{89.4}  & \underline{84.1}  & \underline{88.5}  & \underline{83.8}  & \underline{85.0}  & \underline{85.4} & \underline{91.9} \\
        SANSA~\cite{cuttano2025sansa} & SAM2-L & NIPS’25 & 78.1  & 80.0  & 68.3  & 66.1  & 73.1 & -  & - & - & - & - & - & - \\
        \rowcolor{blue!12} 
        MESAM (Ours) & SAM2-S & -- & \textbf{83.7}  & \textbf{88.9}  & \textbf{83.7}  & \textbf{83.7}  & \textbf{85.0} & \textbf{91.6} & \textbf{84.2}  & \textbf{90.1}  & \textbf{86.2}  & \textbf{87.5}  & \textbf{87.0} & \textbf{92.8} \\ 
        \bottomrule
    \end{tabular}
    }
    \label{tab:pascal}
    \vspace{-5pt}
\end{table*}

\noindent\textbf{Extension from 1-shot to $k$-shot.} The 1-shot pipeline described in Section~\ref{sec:method_pipline} can be easily extended to the $k$-shot setting by modifying the following operations: \textbf{1) Query FG visual prior mask}: We first obtain $k$ different support FG prototypes $Pro^{fg}_s$ by processing each support sample independently, and then use them to calculate $k$ query visual priors through Eq.~(\ref{eq:visual_priors}). Subsequently, these priors are averaged to generate the final query visual prior mask $P_q^v$. \textbf{2) Support adaptive visual prior mask:} We repeat the 1-shot operation for each support sample, yielding $k$ support adaptive visual prior masks $P_s^v$. \textbf{3) BG suppression in ISMR:} Each support adaptive visual memory is matched with the $Q$ to compute a cosine similarity map. The $k$ similarity maps are averaged to obtain the final similarity score $A_{sq}$.

\section{Experiments}
\subsection{Experimental Settings}
\noindent\textbf{Dataset.} In this section, we utilize three datasets with increasing levels of difficulty for our experiments. \textbf{PASCAL-$5^i$} consists of 20 categories from PASCAL VOC 2012 \cite{everingham2010pascal} with additional annotations from SDS \cite{hariharan2011semantic}. \textbf{COCO-$20^i$} is built upon MSCOCO \cite{lin2014microsoft} and contains 80 classes, presenting more challenging due to multiple tiny objects and complex BG scenarios. \textbf{LVIS-$92^i$}~\cite{ICLR2024_4df9a5e6} is constructed based on LVIS~\cite{gupta2019lvis}, which contains 920 fine-grained object categories and significantly higher visual complexity. With these datasets, we evaluate the segmentation performance of our method under two different FSS settings: \textbf{1) For the Standard Setting}, we evenly split the object categories of PASCAL-$5^i$ and COCO-$20^i$ into four disjoint folds, and conduct the experiments in a cross-validation manner (\ie three folds are used for model training and the remaining one for testing with 1,000 episodes sampled). \textbf{2) For the Cross-Dataset Setting}, we examine the generalization ability of MESAM when the training and testing samples are drawn from different data distributions. To this end, we directly evaluate the 1-shot model trained for COCO-$20^0$ (using the standard setting) on each LVIS-$92^i$ fold, where 2,300 episodes are sampled.

\noindent\textbf{Evaluation Metrics.}
Consistent with standard FSS protocols~\cite{zou2025balancing, xu2025unlocking}, two evaluation metrics are used: 1) \textbf{mIoU} measures the overall segmentation quality by averaging IoU across all foreground classes; 2) \textbf{FB-IoU} separately calculates FG and BG IoUs, then takes their average to evaluate class balance performance.

\noindent\textbf{Implementation Details.} Our MESAM model is trained on 4 NVIDIA RTX 4090 (24GB) GPUs and evaluated on a single 4090 GPU. We use the AdamW optimizer to train our ISMR module, as well as SAM2's memory encoder, memory attention module, and mask decoder, while keeping all other components frozen. The learning rate is initialized to 0.001 and scheduled with a polynomial decay strategy~\cite{tian2020prior}. During training, the same data augmentation techniques as in previous works~\cite{xu2025unlocking, cuttano2025sansa} are employed, the input images are randomly cropped into $512\times512$-sized patches. For evaluation, the testing predictions are resized back to the original resolution to compute the mIoU and FB-IoU scores. The batch size is set to 8. The number of training epochs is set to 50 for PASCAL-$5^i$ and 10 for COCO-$20^i$.

\begin{table*}[t]
    \centering
    \caption{Performance comparisons on COCO-$20^i$.``$20^i$'' shows the mIoU scores of 20 novel classes in fold $i$, ``Mean'' denotes the averaged mIoU score across all four folds. The best and second-best results are highlighted in \textbf{bold} and \underline{underlined}, respectively.}
    \vspace{-2pt}
    \resizebox{0.8\textwidth}{!}{
    \begin{tabular}{l|c|c|cccccc|cccccc}
        \toprule
        & & & \multicolumn{6}{c|}{1-shot} & \multicolumn{6}{c}{5-shot} \\
         Methods & Backbone & Venue & $20^0$ & $20^1$ & $20^2$ & $20^3$ & Mean & FB-IoU & $20^0$ & $20^1$ & $20^2$ & $20^3$ & Mean & FB-IoU\\ \midrule
         ABCBNett~\cite{zhu2024addressing} & ResNet50 & CVPR’24 & 44.2  & 54.0  & 52.1  & 49.8  & 50.0 & -  & 50.5  & 59.1  & 57.0  & 53.6  & 55.1 & - \\
        PAHNet~\cite{zou2025balancing} & ResNet50 & ICCV‘25 & 45.2  & 58.2  & 54.4  & 51.2  & 52.3 & 75.2 & 53.0  & 64.1  & 60.8  & 58.7  & 59.2 & 78.1 \\
        OCNet~\cite{wen2025object} & ResNet50 & ICCV‘25 & 45.9  & 56.9  & 52.9  & 50.4  & 51.5 & 73.7  & 52.7  & 63.1  & 57.4  & 54.8  & 57.0 & 76.8 \\ \midrule
        Matcher~\cite{ICLR2024_4df9a5e6} & SAM-H & ICLR'24 & 52.7  & 53.5  & 52.6  & 52.1  & 52.7 & - & 60.1  & 62.7  & 60.9  & 59.2  & 60.7 & - \\
        VRP-SAM~\cite{sun2024vrp} & SAM-H & CVPR'24 & 56.0  & 61.3  & 57.9  & 58.8  & 58.5 & - & 61.4  & 69.4  & 65.9  & 64.9  & 65.4 & - \\
        GF-SAM~\cite{zhang2024bridge} & SAM-H & NIPS’24 & 56.6  & 61.4  & 59.6  & 57.1  & 58.7 & - & 67.1  & 69.4  & 66.0  & 64.8  & 66.8 & - \\
        FCP~\cite{park2025foreground} & SAM-H & AAAI'25 & 46.4  & 56.4  & 55.3  & 51.8  & 52.5 & - & 52.6  & 63.3  & 59.8  & 56.1  & 58.0 & - \\
        FounFSS~\cite{chang2024high} & DINOV2 & Arxiv'24 & 56.8  & 61.0  & \underline{64.2}  & 59.7  & 60.4 & - & - & - & - & - & - & - \\
        FS-DINO~\cite{zhuo2025dinov2} & DINOV2 & Arxiv'25 & 58.6  & \underline{66.0} & 62.3 & \underline{62.6} & \underline{62.4} & -  & 62.1  & 70.2  & \textbf{66.1}  & 63.0  & 65.4 & - \\
        FS-SAM2~\cite{forni2025fs} & SAM2-B & ICIP'25 & 49.3  & 57.6  & 58.3  & 56.1  & 55.3 & - & 55.1  & 60.6  & 58.7  & 59.3  & 58.4 & - \\
        FSSAM~\cite{xu2025unlocking} & SAM2-S & ICML'25 & \underline{59.9}  & 65.6  & 62.1  & 61.6  & 62.3 & \underline{77.3} & \textbf{68.6}  & \underline{74.0}  & 64.5  & \underline{69.9}  & \underline{69.3} & \underline{82.9} \\
        SANSA~\cite{cuttano2025sansa} & SAM2-L & NIPS’25 & 57.7  & 61.7  & 62.8  & 58.5  & 60.2 & - & - & - & - & - & 67.3 & - \\
        \rowcolor{blue!12}
        MESAM (Ours) & SAM2-S  & -- & \textbf{63.9}  & \textbf{70.4}  & \textbf{66.2}  & \textbf{68.6}  & \textbf{67.3} & \textbf{81.3} & \underline{67.2}  & \textbf{74.1}  & \underline{66.0}  & \textbf{73.3}  & \textbf{70.2} & \textbf{84.1}\\
        \bottomrule
    \end{tabular}
    }
    \label{tab:coco}
    \vspace{-5pt}
\end{table*}

\begin{table*}[t]
    \centering
    \caption{Performance comparisons on LVIS-$92^i$. ``$92^i$'' shows the mIoU scores of 92 novel classes in fold $i$, ``Mean'' denotes the averaged mIoU score across all ten folds. The 1-shot results are reported, where the best results are highlighted in \textbf{bold}.}
    \vspace{-2pt}
    \resizebox{0.7\textwidth}{!}{
    \begin{tabular}{l|cccccccccccc}
        \toprule
        Methods & $92^0$ & $92^1$ & $92^2$ & $92^3$ & $92^4$ & $92^5$ & $92^6$ & $92^7$ & $92^8$ & $92^9$ & Mean & FB-IoU \\ \midrule
        Matcher~\cite{ICLR2024_4df9a5e6} & 31.4 & 30.9 & 33.7 & 38.1 & 30.5 & 32.5 & 35.9 & 34.2 & 33.0 & 29.7 & 33.0 & 66.2 \\
        SINE~\cite{liu2024simple} & 28.3 & 31.0 & 31.9 & 34.6 & 30.0 & 31.9 & 32.2 & 33.7 & 30.6 & 27.8 & 31.2 & 63.5 \\
        FSSAM~\cite{xu2025unlocking} & 34.7 & 37.8 & 37.2 & 41.1 & 33.9 & 38.1 & 40.6 & 38.9 & 36.9 & 33.8 & 37.3 & 68.4 \\ \midrule
        MESAM (Ours) & \textbf{38.0}  & \textbf{44.7}  & \textbf{43.5}  & \textbf{44.8}  & \textbf{38.2}  & \textbf{41.4}  & \textbf{44.0}  & \textbf{41.2}  & \textbf{44.0}  & \textbf{39.4}  & \textbf{41.9}  & \textbf{70.9} \\
        \bottomrule
    \end{tabular}
    }
    \label{tab:lvis92}
    \vspace{-8pt}
\end{table*}

\vspace{-3pt}
\subsection{Comparison with State-of-the-Art Methods}
\noindent \textbf{Quantitative Comparison under the Standard Setting.} 
We compare the segmentation performance of our MESAM method with that achieved by recently proposed approaches. Three traditional FSS models built upon the ResNet50 backbone and several foundation-based methods are selected for comparison. The results for the PASCAL-$5^i$ and COCO-$20^i$ datasets under both 1-shot and 5-shot settings are presented in Table~\ref{tab:pascal} and Table~\ref{tab:coco}. We can get the following observations from these tables: \textit{1) Foundation-based methods generally yield better results.} Even the best traditional model (PAHNet) still performs worse than most approaches built upon SAM, DINOv2, and SAM2. We believe this is because these foundation models have already acquired more transferable prior knowledge from large-scale data during their pre-training process. This knowledge and the prompt-based segmentation capability of SAM/SAM2 can both alleviate the learning burden on base classes, thereby reducing the risk of model overfitting to these classes and ultimately benefiting FSS tasks. \textit{2) By constructing the semantic memory, MESAM significantly outperforms other competing baselines, achieving new state-of-the-art results.} Specifically, compared with other SAM2-based models that rely solely on visual cues, MESAM realizes average mIoU score improvements of 4.0\%-11.9\%, 1.6\%-12.2\% 
on PASCAL-$5^i$, and 5.0\%-12.0\%, 0.9\%-11.8\% on COCO-$20^i$ under the two settings, respectively. As for the FB-IoU scores, the performance gains achieved by MESAM over FSSAM are 2.2\% (1-shot) and 0.9\% (5-shot) on PASCAL-$5^i$, which can be more pronounced on COCO-$20^i$, reaching 4.0\% (1-shot) and 1.2\% (5-shot). All these results show the superiority of our MESAM method and the effectiveness of introducing the semantic information to empower SAM2 in addressing FSS tasks.

\noindent \textbf{Quantitative Comparison under the Cross-Dataset Setting.} 
We further evaluate our method on the more challenging LVIS-$92^i$~\cite{gupta2019lvis}, as shown in Table~\ref{tab:lvis92}. MESAM consistently achieves superior performance across all folds, obtaining a mean mIoU of 41.9\%, which surpasses Matcher (33.0\%) and SINE (31.2\%) by 8.9\% and 10.7\%, respectively. Compared with FSSAM, which already exhibits strong cross-dataset generalization, MESAM further improves the performance by 4.6\% (37.3\% $\rightarrow$ 41.9\%). In addition, MESAM achieves the highest FB-IoU score of 70.9\%. These results further validate the effectiveness of introducing semantic information and highlight the strong generalization capability of MESAM.

\begin{figure*}[t]
    \centering
    \includegraphics[width=0.82\textwidth]{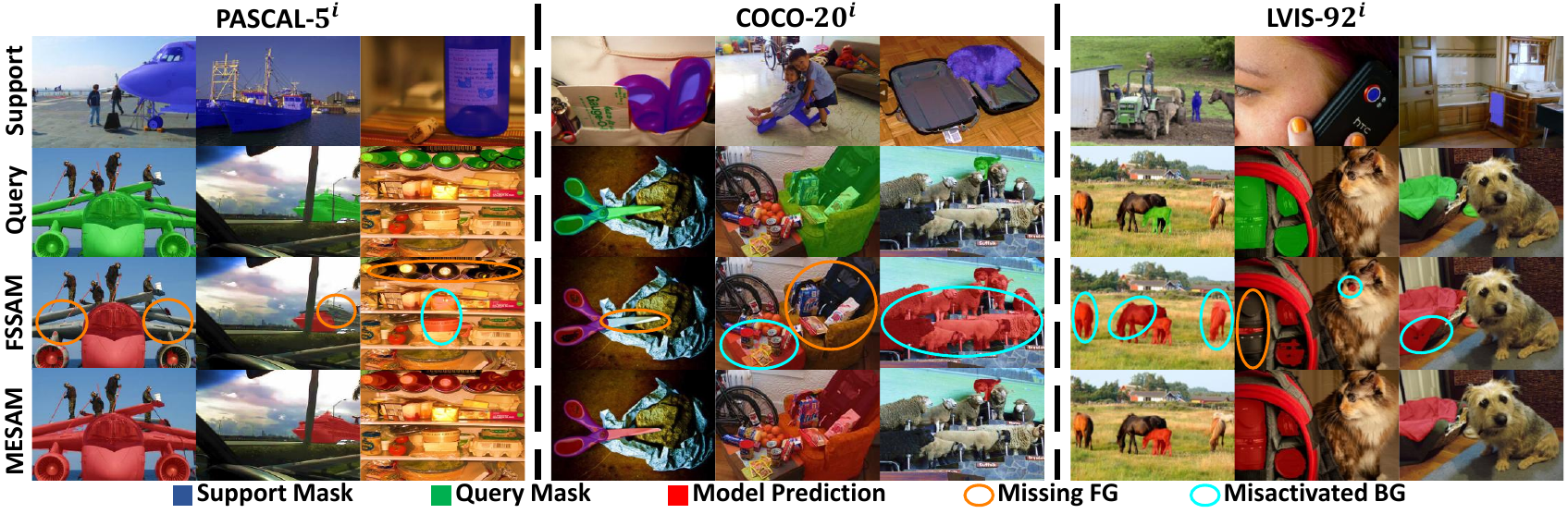}
    \vspace{-5pt}
    \caption{Qualitative comparisons between FSSAM and MESAM on PASCAL-$5^i$, COCO-$20^i$ and LVIS-$92^i$ under the 1-shot setting.}
    \label{fig:final_result}
    \vspace{-10pt}
\end{figure*}

\noindent \textbf{Qualitative Comparison.} To better understand the effectiveness of our method, we visualize the segmentation results of FSSAM (which achieves the second-best performance in most cases) and MESAM in Fig.~\ref{fig:final_result} to make a qualitative comparison. It can be seen that, with large appearance differences between the support and query images, FSSAM often fails to segment the target object completely, and may also produce spurious BG activations. For instance, in the fourth column of the PASCAL-$5^i$ dataset, the support image displays the frontal view of a ``bottle'' (bottle body), while the query image presents the bottles from a top-down view (bottle opening), causing FSSAM to misidentify the target objects and segment irrelevant regions. Similarly, in the second column of the COCO-$20^i$ dataset, the support image only shows the handle of a pair of ``scissors'', resulting in FSSAM missing the scissor blades in the query image. Likewise, in the second column of the LVIS-$92^i$ dataset, the significant appearance gap between the phone ``camera lens'' in the support image and the ``camera lens'' in the query image leads FSSAM to produce incomplete segmentation and mistakenly activate on the eyes of a cat. In contrast, our MESAM method can generate relatively accurate and complete predictions for the target objects in query images, demonstrating its capability to address the challenge of substantial support-query appearance divergence.

\vspace{-3pt}
\subsection{Ablation Studies}
In this section, we investigate how different implementations of our MESAM method affect its segmentation performance. Note that unless otherwise specified, all experiments are conducted on the COCO-$20^i$ dataset using the 1-shot setup.

\begin{table}[t]
    \centering
    \caption{Ablation study on the effect of different memories.}
    \vspace{-2pt}
    \resizebox{0.9\linewidth}{!}{
    \begin{tabular}{l|ccccc}
    \toprule
        Settings & $20^0$ & $20^1$ & $20^2$ & $20^3$ & Mean  \\ \midrule
        A: Only $Mem^s_q$ & 51.2  & 67.1  & 58.4  & 58.7  & 58.9  \\ 
        B: $Mem^s_q + Mem^v_q$ & 62.1 & 68.9 & 65.1  & 66.9 &  65.8 \\
        C: $Mem^s_q + Mem^v_q + Mem^v_s (M_s)$ &  61.9 & 69.9 & 64.7 & 68.1 & 66.2 \\
        D: $Mem^s_q + Mem^v_q + Mem^v_s$ & 63.9  & 70.4  & 66.2  & 68.6  & 67.3  \\
        \bottomrule
    \end{tabular}
    }
    \label{tab:Component-wise}
    \vspace{-10pt}
\end{table}

\noindent \textbf{Effect of Different Memories.} In our SVMG module, we construct three types of memories based on their corresponding prior masks. To study the effect of each memory, we establish our MESAM models under four different settings: ``\textbf{A:} Only $Mem^s_q$'' performs the segmentation based solely on the query semantic memory $Mem^s_q$; ``\textbf{B:} $Mem^s_q + Mem^v_q$'' refines $Mem^s_q$ only through the cross-attention with $Mem^v_q$; ``\textbf{C:} $Mem^s_q + Mem^v_q + Mem^v_s (M_s)$'' calibrates the above refinement process using the support memory $Mem^v_s (M_s)$, which directly takes $M_s$ as the prior mask; ``\textbf{D:} $Mem^s_q + Mem^v_q + Mem^v_s$'' is the full MESAM model. The results obtained by these variations are shown in Table~\ref{tab:Component-wise}. As can be seen, 1) Since the semantic memory alone cannot represent all the FG features in each query image, Setting A achieves the lowest mIoU score of 58.9\%, showing the importance of our ISMR module. 2) Settings C and D both outperform Setting B, suggesting that suppressing the BG noise in attention scores facilitates the memory refinement process. 3) Setting D surpasses Setting C by a large margin of 1.1\%, indicating the effectiveness of our generated support visual prior mask $P_s^v$ in identifying the query-oriented FG features within support images. 

\begin{figure}[t]
    \centering
    \includegraphics[width=0.8\linewidth]{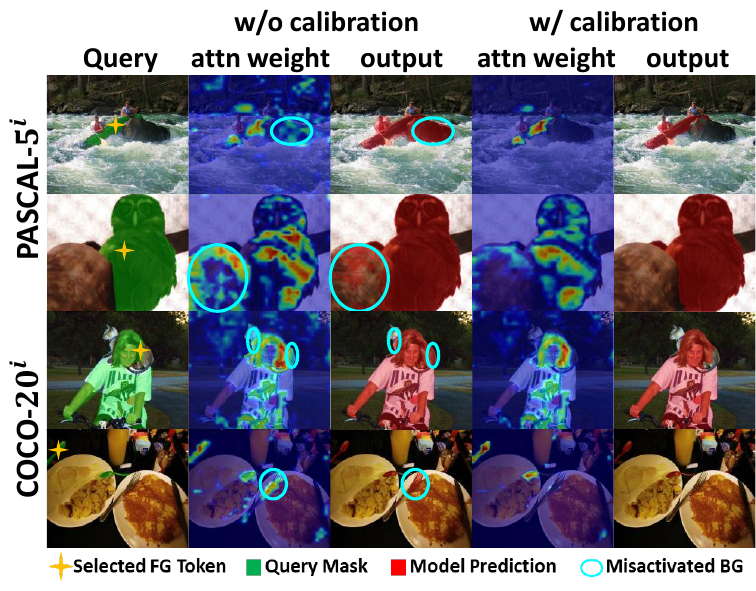}
    \vspace{-2pt}
    \caption{The cross-attention weights and the segmentation results obtained without and with the support calibration.}
    \label{fig:SCMR_attn_weight}
    \vspace{-10pt}
\end{figure}

Figure~\ref{fig:SCMR_attn_weight} visualizes the cross-attention weights for a selected FG pixel (see orange cross) and segmentation results obtained under Settings B and D, \ie w/o and w/ the support calibration. We can find that using pure FG information in $Mem^v_s$ to calibrate attention scores can effectively reduce the influence of misactivated BG areas (see blue circles), thus improving the segmentation performance. 

\begin{table}[t]
    \centering
    \caption{Ablation study on the effect of different ISMR.}
    \vspace{-2pt}
    \resizebox{0.75\linewidth}{!}{
    \begin{tabular}{lccccc}
    \toprule
         & $20^0$ & $20^1$ & $20^2$ & $20^3$ & Mean  \\ \midrule
        Cosine over $K$ & 61.7  & 70.1  & 66.7  & 67.1  & 66.4  \\ 
        FG Enhancement & 62.0 & 69.8 & 67.2  & 67.7 &  66.7 \\
        Memory Swap &  61.1 & 70.5 & 67.8 & 67.9 & 66.8 \\
        ISMR in MESAM & 63.9  & 70.4  & 66.2  & 68.6  & 67.3  \\
        \bottomrule
    \end{tabular}
    }
    \label{tab:Detailed-SCMR}
    \vspace{-10pt}
\end{table}

\noindent \textbf{Effect of Different ISMR Designs.} We also explore the best design of our ISMR module. We modify some operations in ISMR and report the mIoU scores achieved by the modified modules in Table~\ref{tab:Detailed-SCMR}. When the computation of $A_{sq}$ in Eq.~(\ref{eq:cosine_similarity}) is based on the cosine similarity over $K$ instead of $Q$ (``Cosine over $K$''), the overall performance drops from 67.3\% to 66.4\%. This indicates that $Q$ is more effective than $K$ in identifying potential background regions because it derives from more reliable semantic memory. When replacing $\text{min}[\cdot,0]$ in Eq.~(\ref{eq:calibration}) with $\text{max}[\cdot,0]$ to preserve the positive entries in $A_{qq}'$ for enhancing FG information in $A_{qq}''$ (``FG Enhancement''), the negative impact of BG noise remains unmitigated, leading to a 0.6\% decrease in segmentation accuracy. Finally, swapping the roles of semantic and visual memories in the cross-attention (``Memory Swap'') results in a 0.5\% performance degradation, showing the effectiveness of treating the semantic memory as a foundation and enhancing it with visual information.

\section{Conclusion}
This work reveals the superiority of the semantic memory in handling support-query appearance divergences when applying SAM2 to FSS tasks, and explains why it exhibits stronger compatibility with SAM2. With this analysis, we develop our MESAM method primarily upon the semantic memory, which is generated based on the GradCAM of CLIP's text classifier. We then utilize the visual memories derived from the DINOv2 image representations to enhance such semantic memory, enabling it to integrate more fine-grained visual details. Experimental results on PASCAL-$5^i$, COCO-$20^i$, and LVIS-$92^i$ show that MESAM consistently achieves new state-of-the-art performance, demonstrating its effectiveness.

\newpage

\section*{Acknowledgments}
This work was supported by the Young Scientists Fund of the National Natural Science Foundation of China (Grant No. 62306219) and the Project of Sanya Yazhou Bay Science and Technology City (Grant No. SKJC-JYRC-2024-65).

\bibliographystyle{ACM-Reference-Format}
\bibliography{sample-base}
\newpage
\textcolor{white}{1}
\newpage
\appendix

\section{Mask Erosion and Dilation} \label{sec:algorithm}
\subsection{Algorithmic Details}
To analyze the impact of different prior coverage patterns on memory construction in SAM2-based FSS, we conduct a toy experiment that generates pseudo prior masks with varying spatial extents by applying the mask dilation, vanilla-erosion, and skeleton-erosion to the ground-truth segmentation masks of query images. These operations are designed to simulate different statistical behaviors observed in real visual and semantic prior masks.

\begin{algorithm}[h]
\caption{Mask Dilation}
\label{alg:Dilation}
\begin{tabbing}
\hspace{2em}\=\hspace{6em}\=\kill
\textbf{Input:} Query ground-truth mask $M_q$, target ratio $r$, \\ kernel size $k$, padding parameter $pad$\\
\textbf{Output:} Dilated mask $M$\\[0.3em]
\textbf{Stage 1: Initialization}\\
$M \gets M_q$;\\
$kernal \gets$ ones($k, k$);\\
$initial\_area \gets$ sum of non-zero pixels in $M$;\\
$target\_area \gets initial\_area \times (1 + r)$;\\
$current\_area \gets initial\_area$;\\
$step \gets 0$;\\
\textbf{Stage 2: Iterative Dilation} \\
\textbf{while} $current\_area < target\_area$ \textbf{do}\\
1: \> $M \gets \mathbb{I}(\text{Conv2D}(M, kernal, pad)> 0)$;\\
2: \> $current\_area \gets \text{sum}(\mathbb{I}(M > 0))$;\\
3: \> $step \gets step + 1$;\\
4: \> \textbf{if} $step > 1000$: \textit{then break;}\\
$M \gets \mathbb{I}(M > 0)$
\end{tabbing}
\end{algorithm}

\begin{figure*}[t]
    \centering
    \includegraphics[width=1\linewidth]{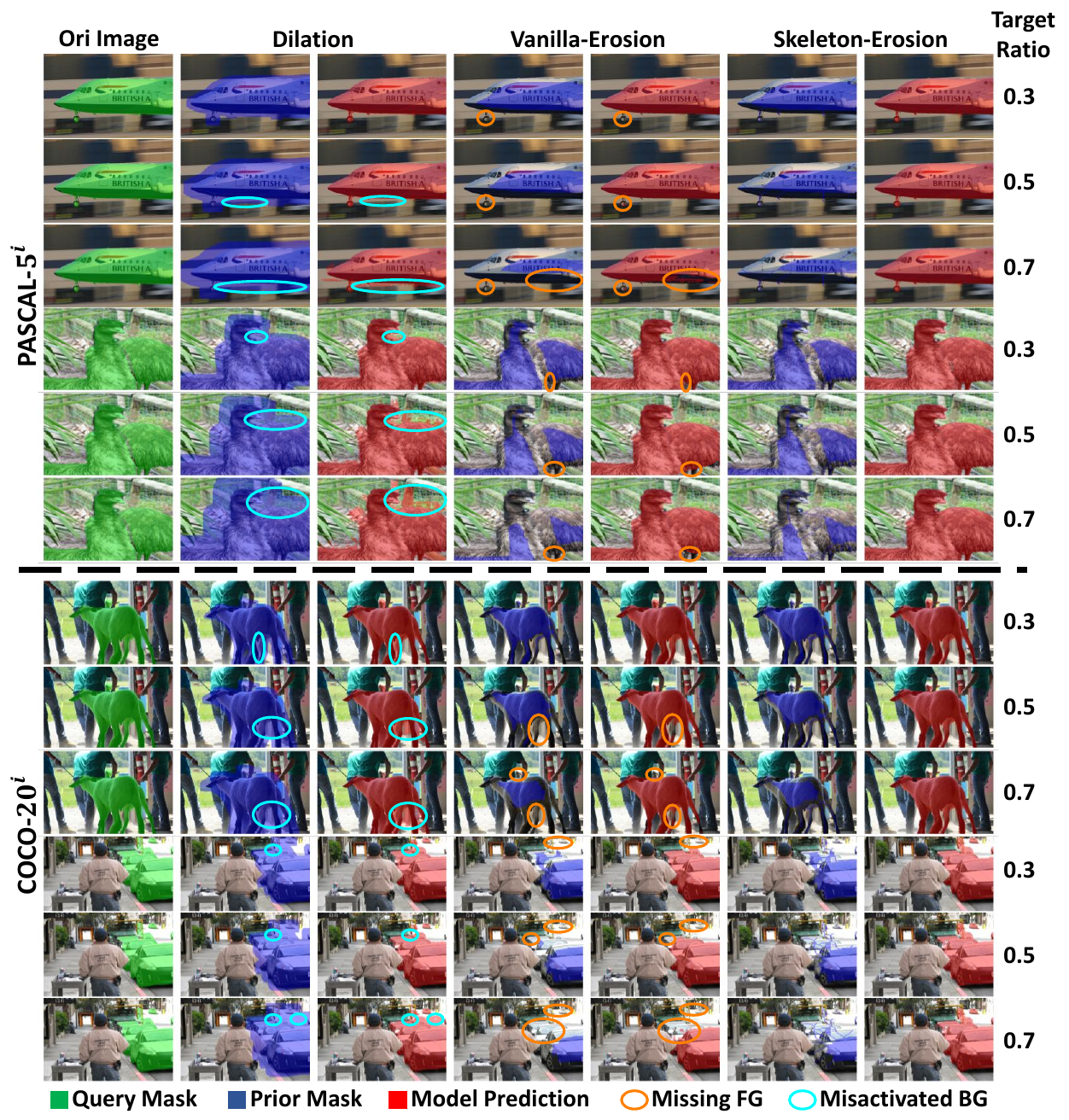}
    \caption{Qualitative comparisons between dilation, vanilla-erosion, and skeleton-erosion operations with varying target ratios.}
    \label{fig:erode_dilate}
\end{figure*}

\begin{figure*}[t]
    \centering
    \includegraphics[width=1\linewidth]{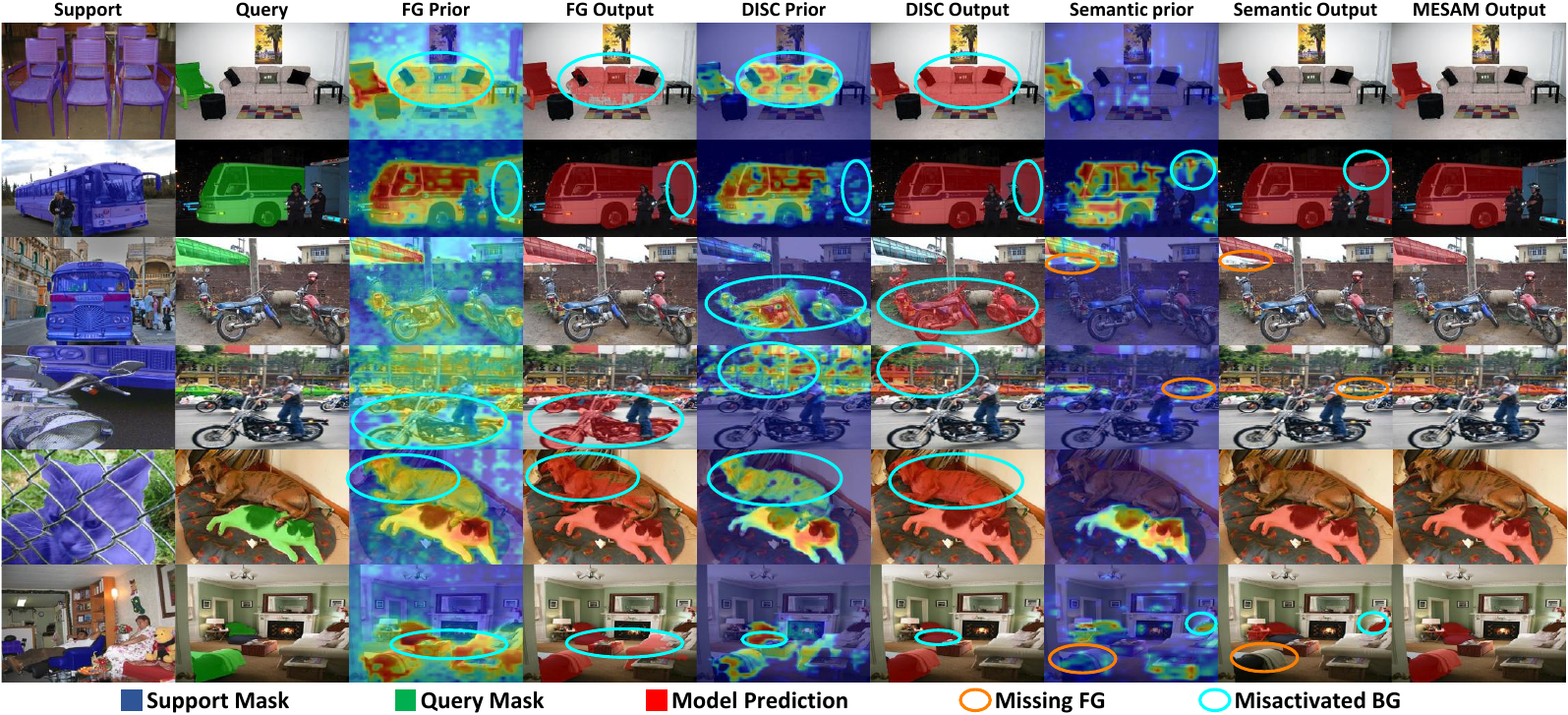}
    \caption{The prior masks and the segmentation results of visual memories (FG, DISC) and semantic memory on PASCAL-$5^i$.}
    \label{fig:memory_pascal}
\end{figure*}
\begin{figure*}[t]
    \centering
    \includegraphics[width=1\linewidth]{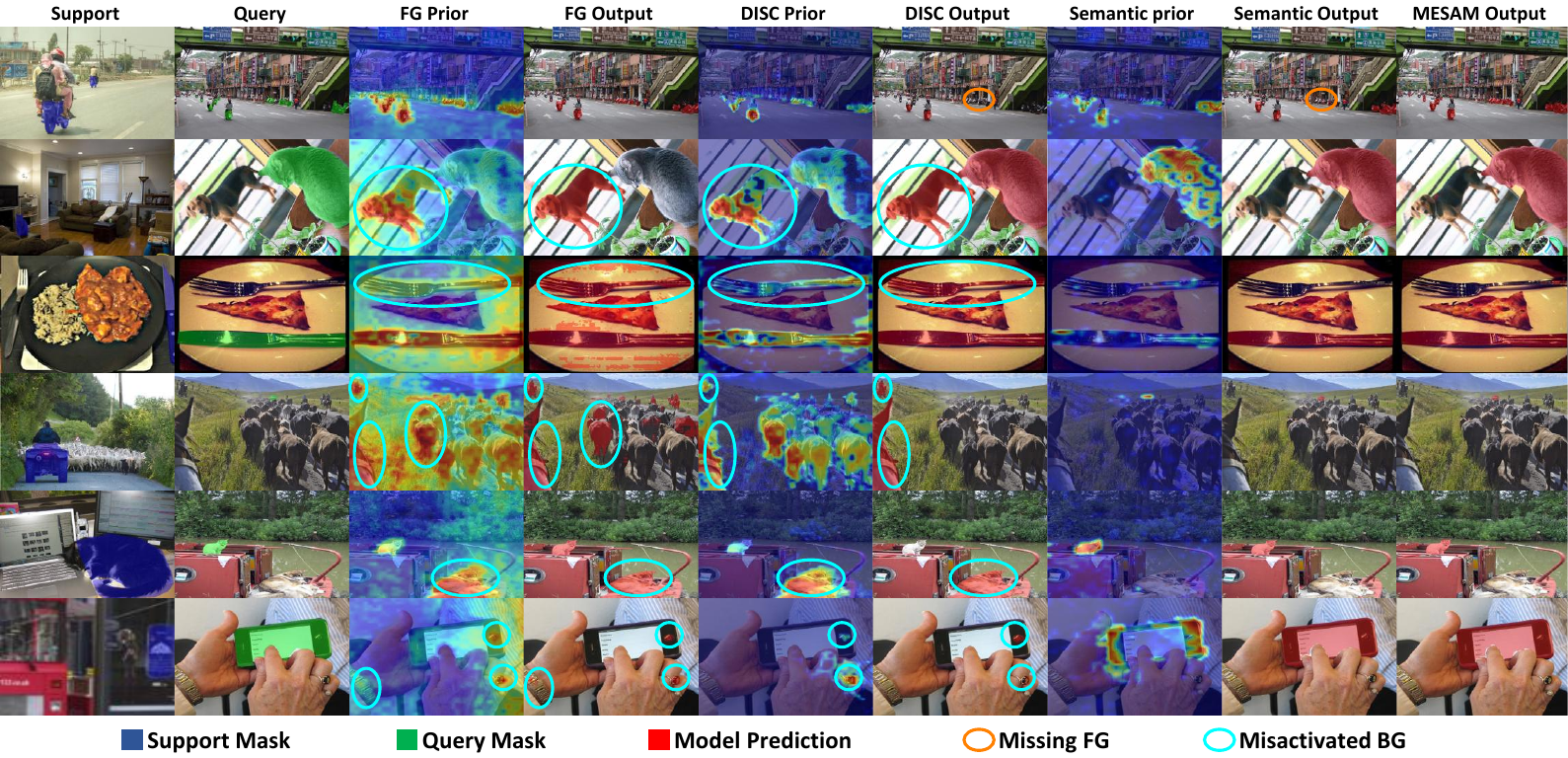}
    \caption{The prior masks and the segmentation results of visual memories (FG, DISC) and semantic memory on COCO-$20^i$.}
    \label{fig:memory_coco}
\end{figure*}

\begin{algorithm}[h]
\caption{Mask Erosion}
\begin{tabbing}
\hspace{2em}\=\hspace{6em}\=\kill
\textbf{Input:} Query ground-truth mask $M_q$, target ratio $r$, \\ kernel size $k$, padding parameter $pad$\\
\textbf{Output:} Eroded mask $M$\\[0.3em]
\textbf{Stage 1: Initialization}\\
$M \gets M_q$;\\
$kernal \gets$ ones($k, k$);\\
$initial\_area \gets$ sum of non-zero pixels in $M$;\\
$target\_area \gets initial\_area \times (1 - r)$;\\
$current\_area \gets initial\_area$;\\
$step \gets 0$;\\
$S \gets \text{Skeleton}(M);$ \quad \text{// Generating skeleton from mask}\\[0.3em]
\textbf{Stage 2: Iterative Erosion (Two implementations)}\\[0.3em]
\textbf{Option 2a: Vanilla-Erosion} \\
\textbf{while} $current\_area > target\_area$ \textbf{do}\\
1: \> $M \gets 1-\mathbb{I}(\text{Conv2D}(1-M, kernal, pad)> 0)$;\\
2: \> $current\_area \gets \text{sum}(\mathbb{I}(M > 0))$;\\
3: \> $step \gets step + 1$;\\
4: \> \textbf{if} $step > 1000$: \textit{then break;}\\
$M \gets \mathbb{I}(M > 0)$\\[0.3em]
\textbf{Option 2b: Skeleton-Erosion} \\
\textbf{while} $current\_area > target\_area$ \textbf{do}\\
1: \> $M \gets 1-\mathbb{I}(\text{Conv2D}(1-M, kernal, pad)> 0)$;\\
2: \> $current\_area \gets \text{sum}(\mathbb{I}(M+S > 0))$;\\
3: \> $step \gets step + 1$;\\
4: \> \textbf{if} $step > 1000$: \textit{then break;}\\
$M \gets \mathbb{I}(M + S > 0)$
\end{tabbing}
\label{alg:Erosion}
\end{algorithm}

\noindent \textbf{Dilation.}
As shown in Algorithm~\ref{alg:Dilation}, starting from the query ground-truth mask $M_q$, the dilation operation repeatedly performs a 2D convolution (with a $3 \times 3$ all-ones kernel and a padding parameter $pad$) on the mask $M$ obtained in the previous iteration, until the area of its FG regions becomes $1+r$ times larger than that in $M_q$. This process typically introduces additional BG pixels, similar to the FG visual prior mask with higher FP rates. 

\noindent \textbf{Vanilla-Erosion.}
The only difference between the vanilla-erosion and dilation operations lies in Step 1 of Option~2a within Algorithm~\ref{alg:Erosion}, where the 2D convolution is applied to $1-M$ and the resulting mask is inverted to obtain the current $M$. The iteration terminates when the FG area of $M$ is $1-r$ times smaller than that in $M_q$. The mask generated by this operation will exhibit high reliability because the BG regions are filtered out. However, it will also result in the loss of certain important FG information, thus leading to higher FN rates, similar to the DISC visual prior mask.

\noindent \textbf{Skeleton-Erosion.}
As shown in Option~2b of Algorithm~\ref{alg:Erosion}, the FG regions obtained by the skeleton-erosion operation are defined as the union area of the eroded mask $M$ and the skeleton $S$ (see $M+S$ in Step 2). By doing so, this process further extends the vanilla-erosion by preserving the representative skeletal structure of target objects in $M_q$ unchanged. Consequently, the resulting mask $M$ is similar to our semantic prior mask, which demonstrates both high reliability and representativeness.

\subsection{Qualitative Analysis}
In Fig.~\ref{fig:erode_dilate}, we visualize the prior masks generated by these three operations under different target ratios and their corresponding segmentation results. As can be seen, the dilation operation introduces BG noise into the prior mask, resulting in erroneous activations within BG regions (see blue circles). Meanwhile, the vanilla-erosion removes some FG details, leading to incomplete segmentation of targets or the omission of certain tiny objects (see orange circles). In contrast, by retaining the skeleton $S$, the skeleton-erosion yields much better predictions than the other two operations, even though its prior mask is quite similar to that produced by the vanilla-erosion. These results indicate that generating prior masks with reliable and representative FG information is crucial for achieving precise FSS segmentation with SAM2. This also explains why our semantic memory is compatible with SAM2, as it introduces classification-oriented semantic information that demonstrates both reliability and representativeness for the target category.

As a supplement to the main text, we present more prior masks and segmentation results of two visual memories (\ie FG and DISC) and our semantic memory in Fig.~\ref{fig:memory_pascal} and Fig.~\ref{fig:memory_coco}. This provides additional evidence to support our claim: \textit{The semantic memory outperforms visual memories in solving FSS tasks. But it cannot cover all FG details in query images by itself, so it is necessary to refine it with fine-grained visual information.} Such insight motivates the design of our MESAM method.

\begin{table*}[t]
    \centering
    \caption{Performance of MESAM on PASCAL-$5^i$ with five random seeds \{321 (default), 0, 1, 2, 3\}. 1,000 episodes are randomly sampled for evaluation. ``Average'' and ``Std'' denote the mean mIoU scores and the standard deviations across different random seeds.}
    \resizebox{0.75\textwidth}{!}{
    \begin{tabular}{c|ccccc|ccccc}
        \toprule
         & \multicolumn{5}{c|}{1-shot} & \multicolumn{5}{c}{5-shot} \\
         Seed & $5^0$ & $5^1$ & $5^2$ & $5^3$ & Mean & $5^0$ & $5^1$ & $5^2$ & $5^3$ & Mean \\ \midrule
         default & 83.70  & 88.93  & 83.74  & {83.70}  & 85.02  & {84.21}  & {90.08}  & 86.21  & 87.50  & {87.00}  \\  
         0 & {83.75}  & 89.29  & {83.88}  & 83.38  & {85.08}  & 83.65  & 89.57  & 86.49  & 87.69  & 86.85  \\
        1 & 83.44  & 88.75  & 83.47  & 83.46  & 84.78  & 83.95  & 89.91  & 86.38  & 87.45  & 86.92  \\ 
        2 & 83.43  & 89.18  & 83.38  & 83.30  & 84.82  & 83.92  & 89.77  & 86.03  & {87.87}  & 86.90  \\ 
        3 & 83.38  & {89.23}  & 83.82  & 83.33  & 84.94  & 84.11  & 89.66  & {86.62}  & 87.15  & 86.89  \\ \midrule
        Average & 83.54  & 89.08  & 83.66  & 83.43  & 84.93  & 83.97  & 89.80  & 86.35  & 87.53  & 86.91  \\ 
        Std & 0.17  & 0.23  & 0.22  & 0.16  & 0.13  & 0.21  & 0.20  & 0.23  & 0.27  & 0.06 \\
        \bottomrule
    \end{tabular}
    }
    \label{tab:pascal_seed}
\end{table*}

\begin{table*}[t]
    \centering
    \caption{Performance of MESAM on COCO-$20^i$ with five random seeds \{321 (default), 0, 1, 2, 3\}. 1,000 episodes are randomly sampled for evaluation. ``Average'' and ``Std'' denote the mean mIoU scores and the standard deviations across different random seeds.}
    \resizebox{0.75\textwidth}{!}{
    \begin{tabular}{c|ccccc|ccccc}
        \toprule
         & \multicolumn{5}{c|}{1-shot} & \multicolumn{5}{c}{5-shot} \\
         Seed & $20^0$ & $20^1$ & $20^2$ & $20^3$ & Mean & $20^0$ & $20^1$ & $20^2$ & $20^3$ & Mean \\ \midrule
         default & {63.92} & 70.41 & {66.17} & {68.63} & {67.28}  & 67.22 & 74.13 & 66.01 & {73.32} & 70.17  \\ 
          0 & 62.97 & 70.59 & 65.13 & 67.48 & 66.54  & 65.59 & 72.84 & {68.16} & 72.64 & 69.81  \\
        1 & 63.55 & 70.33 & 65.46 & 67.21 & 66.64  & {68.52} & {74.45} & 63.31 & 72.79 & 69.77  \\
        2 & 63.38 & 70.12 & 62.33 & 67.59 & 65.86  & 68.16 & 73.45 & 66.96 & 72.70 & {70.32}  \\
        3 & 62.65 & {70.60} & 64.94 & 68.07 & 66.57  & 65.61 & 73.65 & 67.25 & 71.85 & 69.59  \\
         \midrule
        Average & 63.29  & 70.41  & 64.81  & 67.80  & 66.58  & 67.02  & 73.70  & 66.34  & 72.66  & 69.93  \\ 
         Std & 0.50  & 0.20  & 1.46  & 0.56  & 0.51  & 1.38  & 0.62  & 1.86  & 0.53  & 0.30 \\
        \bottomrule
    \end{tabular}
    }
    \label{tab:coco_seed}
\end{table*}

\begin{table*}[t]
    \centering
    \caption{Performance of MESAM on COCO-$20^i$ evaluated with varying numbers of testing episodes \{1,000, 4,000, 10,000, 20,000\}. “Time” denotes the total inference time for the evaluation process. ``Average'' and ``Std'' denote the mean mIoU scores and the standard deviations across different numbers of testing episodes.}
    \resizebox{0.95\textwidth}{!}{
    \begin{tabular}{c|cccccc|cccccc}
        \toprule
         & \multicolumn{6}{c|}{1-shot} & \multicolumn{6}{c}{5-shot} \\
         \#Episodes & $20^0$ & $20^1$ & $20^2$ & $20^3$ & Mean &Time (s) & $20^0$ & $20^1$ & $20^2$ & $20^3$ & Mean &Time (s)  \\ \midrule
         1000 & 63.92  & 70.41  & 66.17  & {68.63}  & 67.28  & 143.1  & {67.22} & {74.13} & 66.01 & {73.32} & 70.17  & 218.7  \\
        4000 & {65.38}  & 69.86  & {68.68}  & 66.57  & {67.62}  & 568.2  & 66.75 & 73.81 & {72.52} & 72.32 & {71.35}  & 881.4  \\
        10000 & 63.66  & 70.35  & 67.27  & 65.43  & 66.68  & 1406.5  & 66.38 & 73.73 & 70.73 & 71.84 & 70.67  & 2183.2  \\
        20000 & 63.80  & {70.55}  & 67.41  & 65.54  & 66.83  & 2799.9  & 66.57 & 73.87 & 70.69 & 71.87 & 70.75  & 4425.7  \\ \midrule
        Average & 64.19  & 70.29  & 67.38  & 66.54  & 67.10  & 1229.4  & 66.73  & 73.89  & 69.99  & 72.34  & 70.74  & 1927.3  \\
        Std & 0.80  & 0.30  & 1.03  & 1.48  & 0.43  & -  & 0.3  & 0.17  & 2.79  & 0.69  & 0.48  & - \\
        \bottomrule
    \end{tabular}
    }
    \label{tab:coco_num}
\end{table*}

\section{Additional Experimental Results}
\subsection{Performance Stability}
We repeat the testing procedure of our MESAM method five times using different random seeds \{321 (default), 0, 1, 2, 3\} to investigate its performance stability. Note that these seeds are used only to determine the 1,000 testing episodes by selecting different support and query images from PASCAL-$5^i$ and COCO-$20^i$. The training process is conducted under the default setting, where the seed value is fixed to 321. From Table~\ref{tab:pascal_seed} and Table~\ref{tab:coco_seed}, we can observe that: 1) On PASCAL-$5^i$, the mIoU scores averaged over five seeds are 84.93\% ± 0.13\% (1-shot) and 86.91\% ± 0.06\% (5-shot), showing high consistency with the results reported in the main text achieved under the default setting. The small standard deviations indicate that MESAM demonstrates strong robustness to the randomness in this dataset. 2) On COCO-$20^i$, MESAM achieves results of 66.58\% ± 0.51\% (1-shot) and 69.93\% ± 0.30\% (5-shot), whose standard deviations are larger than those on PASCAL-$5^i$, indicating that the performance on this dataset is more sensitive to episode sampling. It is mainly because the images in COCO-$20^i$ often contain multiple objects, small-sized targets, and complex background scenes. This makes the episodes sampled by different seeds vary significantly in segmentation difficulty, ultimately leading to performance fluctuations. Nevertheless, MESAM with different random seeds still consistently outperforms other competing methods.

% \subsection{Effect of Different Number of Testing Episodes}
As previously discussed, owing to the complexity of the COCO-$20^i$ dataset, using only 1,000 episodes may lead to biased or unstable evaluation of model performance. Therefore, we vary the number of testing episodes within the range of \{1,000 (default), 4,000, 10,000, 20,000\} and summarize the results in Table~\ref{tab:coco_num} to analyze how MESAM performs on more episodes. It can be observed that the mIoU scores obtained with 1,000 episodes exhibit no significant difference compared to those averaged across 1,000, 4,000, 10,000, and 20,000 episodes. Specifically, under 1-shot and 5-shot settings, using 1,000 episodes for evaluation yields mIoU scores of 67.28\% and 70.17\%, respectively, while the average scores are 67.10\% and 70.74\%. The differences between these two are merely 0.18\% and 0.57\%, showing the stability of our MESAM method.

\begin{figure}[t]
    \centering
    \includegraphics[width=1\linewidth]{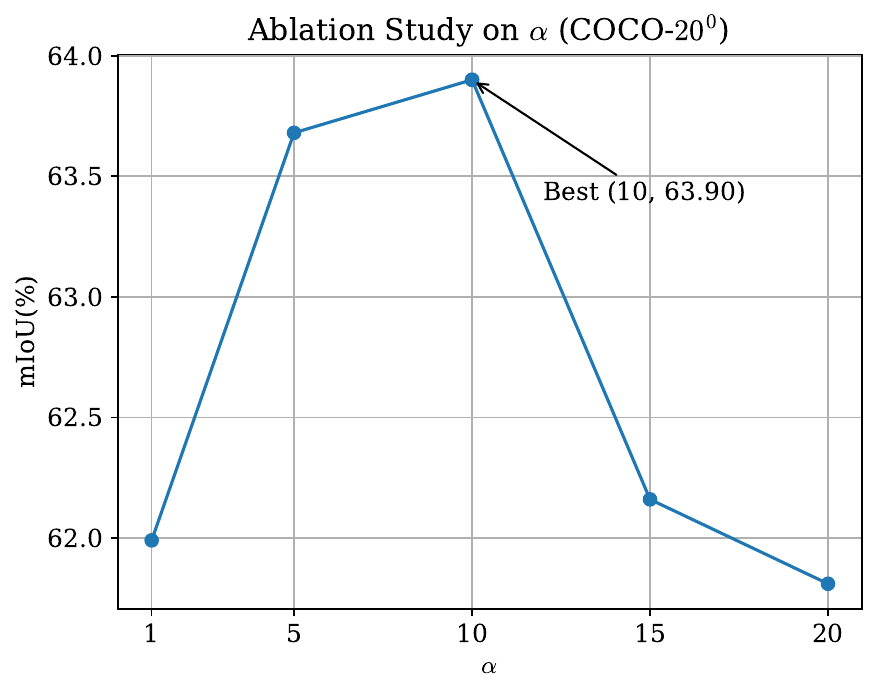}
    \caption{Ablation study of the scaling factor $\alpha$ on COCO-$20^0$.}
    \label{fig:alpha_ablation}
\end{figure}

\begin{table}[t]
    \centering
    \caption{Performance comparisons between different MESAM models built upon four SAM2 versions.}
    \resizebox{1.0\linewidth}{!}{
    \begin{tabular}{l|crrrrr}
    \toprule
        Backbones &Size (M) & $20^0$ & $20^1$ & $20^2$ & $20^3$ & Mean  \\ \midrule
        SAM2-T &38.9 & 62.8  & 70.6  & 66.4  & 69.4  & 67.3   \\
        SAM2-S &46.0 & 63.9  & 70.4  & 66.2  & 68.6  & 67.3 \\
        SAM2-B &80.8 & 63.0  & 69.8  & 65.6  & 67.7  & 66.5 \\
        SAM2-L &224.4 & 62.4  & 71.1  & 68.7  & 68.4  & 67.7 \\
        \bottomrule
    \end{tabular}
    }
    \label{tab:SAM-version}
\end{table}

\subsection{Inference Time Analysis}
In Table~\ref{tab:coco_num}, we also report the inference time (s) of MESAM under each episode configuration. The evaluations are conducted on a single NVIDIA RTX 4090 GPU with 24GB memory, and the SAM2-S model is used as our backbone. 
Although the deployment of large foundation models inevitably introduces additional computational overhead, the inference time of our method is still acceptable. Concretely, MESAM achieves processing speeds of around 7.1 FPS (1-shot) and 4.6 FPS (5-shot), which are comparable to FSSAM~\cite{xu2025unlocking} (7.3 FPS and 4.6 FPS). Meanwhile, MESAM achieves significant performance gains of 5.0\% and 0.9\% over FSSAM under the two settings. This demonstrates that our method maintains a favorable trade-off between segmentation accuracy and model efficiency, whilst benefiting from strong foundation models.

\subsection{Ablation Study on $\alpha$}
We further analyze the impact of the scaling factor $\alpha$ in the attention calibration (Eq. (11)). As shown in Fig.~\ref{fig:alpha_ablation}, the performance first improves as $\alpha$ increases from 1 to 10, with mIoU rising from 61.99 to a peak of 63.90. This suggests that properly amplifying the negative components in $A'_{qq}$ helps suppress BG activations and enhances the quality of attention calibration. However, when $\alpha$ continues to increase beyond 10, the performance gradually declines (62.16 at $\alpha$=15 and 61.81 at $\alpha$=20), indicating that excessively large $\alpha$ may over-amplify negative responses, leading to biased attention scores and degraded segmentation quality. Overall, $\alpha$=10 achieves the best trade-off between effective BG suppression and stable attention calibration, and is adopted in our final model.

\subsection{Effect of Different SAM2 Sizes}
We then investigate the effect of different SAM2 sizes on the performance of our method. We build different MESAM models upon four SAM2 versions (including tiny (T), small (S), base (B), and large (L)) and report their segmentation accuracy in Table~\ref{tab:SAM-version}. It can be found that SAM2-T and SAM2-S achieve comparable mIoU scores (both 67.3\%), while SAM2-B performs slightly worse (66.5\%). Moreover, although SAM2-L yields the best results, its improvements over other smaller variants are marginal, particularly considering its significantly larger parameter size (224.4 M). These results indicate that the effectiveness of our method does not rely on large SAM2 backbones and can achieve strong performance with a relatively lightweight SAM2 version (\ie SAM2-S).  This is desirable for practical deployment when computational resources are limited.

\begin{figure*}[t]
    \centering
    \includegraphics[width=0.95\linewidth]{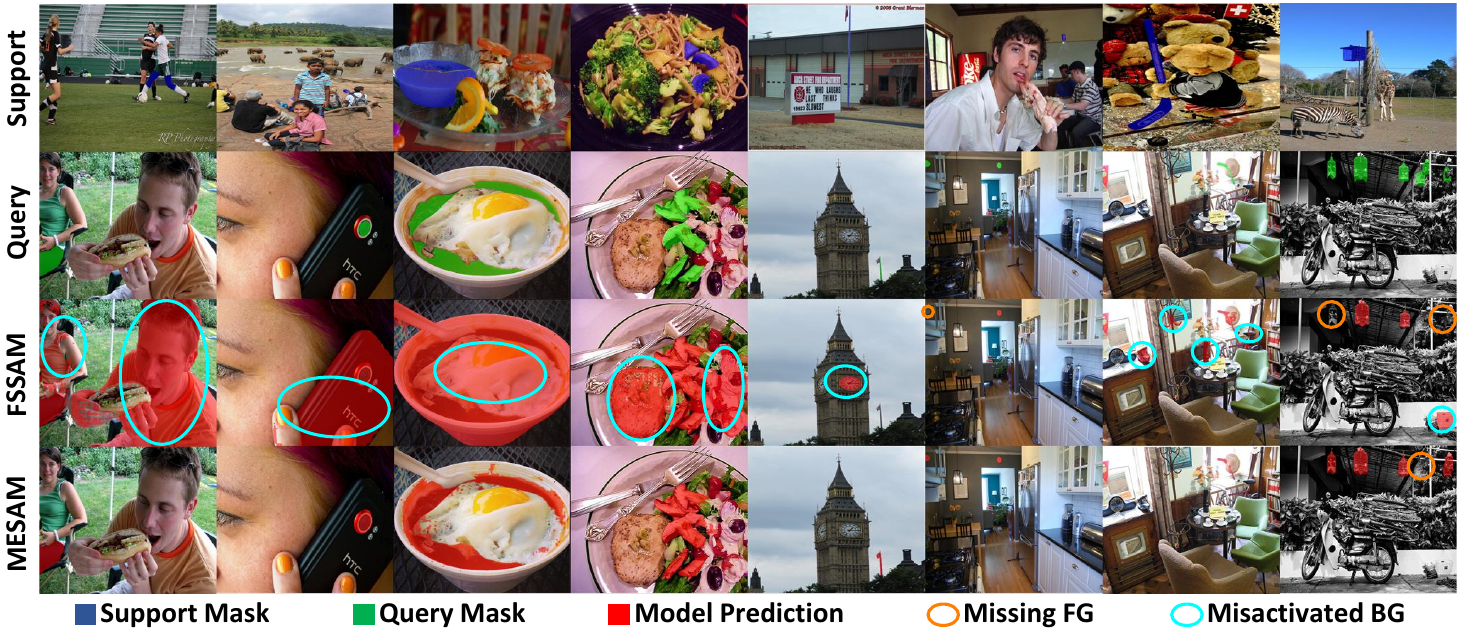}
    \caption{Additional qualitative comparisons between FSSAM and MESAM on LVIS-$92^i$ under the 1-shot setting.}
    \label{fig:lvis_resualt}
\end{figure*}

\begin{table}[t]
    \centering
    \caption{Performance comparisons under COCO-$20^i$$\rightarrow$ PASCAL-$5^i$ cross-dataset setting. The 1-shot results are reported, where the best results are highlighted in \textbf{bold}.}
    \resizebox{1.0\linewidth}{!}{
    \begin{tabular}{l|ccccc}
    \toprule
        Methods & $5^0$ & $5^1$ & $5^2$ & $5^3$ & Mean  \\ \midrule
        HSNet~\cite{min2021hypercorrelation} & 47.0  & 65.2  & 67.1  & 77.1  & 64.1  \\
        VAT~\cite{hong2022cost}& 68.3  & 64.9  & 67.5  & 79.8  & 65.1  \\
        PMNet~\cite{chen2024pixel} & 71.0  & 72.3  & 66.6  & 63.8  & 68.4  \\
        GF-SAM~\cite{zhang2024bridge} & 71.1  & 75.7  & 69.2  & 73.3  & 72.1  \\
        SANSA~\cite{cuttano2025sansa} & 67.1  & 70.9  & 73.5  & 78.6  & 72.5  \\
        FSSAM~\cite{xu2025unlocking} &79.5 & 85.7 & 85.4 & 76.5 & 81.8 \\
        MESAM (ours) & \textbf{83.0}  & \textbf{89.8}  & \textbf{88.2}  & \textbf{81.9}  & \textbf{85.7} \\
        \bottomrule
    \end{tabular}
    }
    \label{tab:COCOtoPascal}
\end{table}

\subsection{Cross-Dataset Generalization}
In this subsection, following HSNet~\cite{min2021hypercorrelation}, we compare the cross-dataset segmentation performance of different methods under a COCO-$20^i$$\rightarrow$PASCAL-$5^i$ setting, where models are trained on COCO-$20^i$ and directly tested on PASCAL-$5^i$. The object categories in the training and testing sets are disjoint. As shown in Table~\ref{tab:COCOtoPascal}, MESAM achieves substantial performance gains across all four folds, outperforming SANSA by 13.2\% and FSSAM by 3.9\% on average, indicating its superior generalization capability under data distribution shifts. Moreover, compared to Table~3 in the main text, the performances of foundation-based approaches (\ie GF-SAM, SANSA, FSSAM, and MESAM) are not significantly degraded, where FSSAM and MESAM even produce better results. It may be because the knowledge learned from the more complicated COCO-$20^i$ dataset can be effectively applied to the relatively easier PASCAL-$5^i$ dataset.

\section{Additional Qualitative Comparison}
\begin{figure*}[t]
    \centering
    \includegraphics[width=0.95\linewidth]{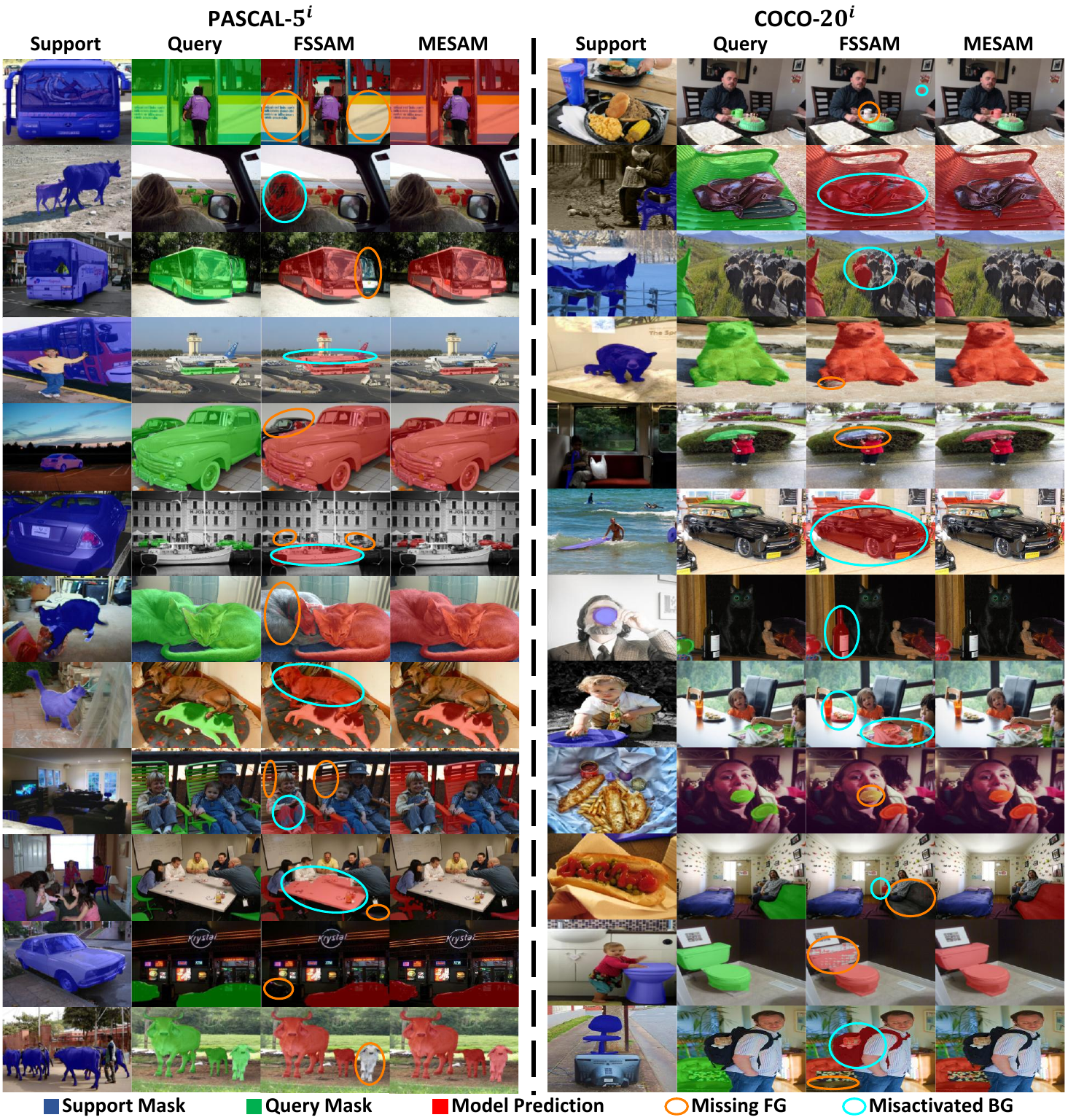}
    \caption{Additional qualitative comparisons between FSSAM and MESAM on PASCAL-$5^i$ and COCO-$20^i$ under the 1-shot setting.}
    \label{fig:additional_final_resualt}
\end{figure*}

Fig.~\ref{fig:additional_final_resualt} provides additional qualitative comparisons between FSSAM and MESAM. Consistent with the main text, FSSAM fails to handle the significant appearance differences between support and query images, leading to incomplete segmentation (e.g., the ``Bus'' and ``Cup'' in the first row) and erroneous BG activations (e.g., the human head and bags in the second row). Furthermore, we present additional cross-dataset qualitative results on the LVIS-$92^i$ dataset. As shown in the last row of Fig.~\ref{fig:lvis_resualt}, FSSAM fails to correctly segment the ``birdcage'' due to appearance discrepancy, leading to missing regions and erroneous activations. By contrast, the predictions yielded by MESAM are more complete and accurate, illustrating that introducing the semantic memory can effectively enhance MESAM's robustness against support-query appearance divergence.

\section{Limitations and Future Works.}
By incorporating the effective memory mechanism we proposed, MESAM realizes consistent improvement over existing FSS approaches. However, as described in the main text, we process multiple support samples by simply averaging their derived prior masks and cosine-similarity maps. These operations may fail to fully leverage the diverse visual information contained in different support images. As a result, our method achieves relatively less significant performance gains under the 5-shot setting. Therefore, in future work, we will explore more effective multi-support fusion mechanisms and adaptive memory selection strategies, enabling the model to better benefit from additional annotations. Moreover, MESAM still requires a training process to update its parameters before it can adapt to the target dataset, which can increase the computational overhead. This motivates us to design more efficient training-free or test-time adaptation strategies, which we will pursue as a future research direction.

\end{document}